\pdfoutput=1

\documentclass[11pt]{article}

\usepackage[final]{acl}

\usepackage{times}
\usepackage{latexsym}

\usepackage[T1]{fontenc}

\usepackage[utf8]{inputenc}

\usepackage{microtype}

\usepackage{inconsolata}
\usepackage{enumitem}

\usepackage{graphicx}

\usepackage{colortbl}
\usepackage{multirow}
\usepackage{xcolor}
\usepackage{tcolorbox} 
\usepackage{booktabs}
\usepackage{subfig}
\definecolor{up}{HTML}{C40000}
\definecolor{down}{HTML}{000000}
\usepackage{lipsum}

\usepackage{ulem}

\title{GeoGPT4V: Towards Geometric Multi-modal Large Language Models with Geometric Image Generation}

\author{
    \textbf{Shihao Cai\textsuperscript{1}} \footnotemark[1] \footnotemark[3],
    \textbf{Keqin Bao\textsuperscript{1}} \footnotemark[1],
    \textbf{Hangyu Guo\textsuperscript{2}},
    \textbf{Jizhi Zhang\textsuperscript{1}},
    \\
    \textbf{Jun Song\textsuperscript{2}} \footnotemark[2],
    \textbf{Bo Zheng\textsuperscript{2}}
    \\
    \\
    \textsuperscript{1}University of Science and Technology of China,
    \textsuperscript{2}Alibaba Group
    \\
    \small{
        \{caishihao, baokq, cdzhangjizhi\}@mail.ustc.edu.cn,
    }
    \\
    \small{hyguo0220@gmail.com, \{jsong.sj, bozheng\}@alibaba-inc.com}
}

\begin{document}
\maketitle

\newcommand\blfootnote[1]{%
\begingroup
\renewcommand\thefootnote{}\footnote{#1}%
\addtocounter{footnote}{-1}%
\endgroup
}

\blfootnote{*These authors contributed equally to this work.}
\blfootnote{\dag Corresponding author.}
\blfootnote{\ddag This work is done when Shihao Cai is an intern at Alibaba.}

\begin{abstract}
Large language models have seen widespread adoption in math problem-solving.
However, in geometry problems that usually require visual aids for better understanding, even the most advanced multi-modal models currently still face challenges in effectively using image information.
High-quality data is crucial for enhancing the geometric capabilities of multi-modal models, yet existing open-source datasets and related efforts are either too challenging for direct model learning or suffer from misalignment between text and images. 
To overcome this issue, we introduce a novel pipeline that leverages GPT-4 and GPT-4V to generate relatively basic geometry problems with aligned text and images, facilitating model learning. 
We have produced a dataset of 4.9K geometry problems and combined it with 19K open-source data to form our GeoGPT4V dataset.
Experimental results demonstrate that the GeoGPT4V dataset significantly improves the geometry performance of various models on the MathVista and MathVision benchmarks.
The code is available at \url{https://github.com/Lanyu0303/GeoGPT4V_Project}.

\end{abstract}
\section{Introduction} 

With large language models (LLMs) demonstrating formidable performance, their application in solving mathematical problems has become an increasingly popular trend~\cite{openmath,DBLP:journals/corr/abs-2312-08935,tora, mathcoder}.
Prior research has indicated that humans encounter a significant reduction in accuracy when resolving geometric problems devoid of visual aids~\cite{dblp:conf/acl/chentqllxl21}.
Thus, the integration of visual information from images is imperative for accurately solving of such mathematical problems, necessitating the visual perception capabilities of multi-modal large language models (MLLMs).
However, even the best batch of MLLMs available now (such as GPT-4V~\cite{openai2023gpt4v}, Gemini~\cite{geminiteam2023gemini}) still lag significantly behind human performance~\cite{wang2024measuring}.
Therefore, researchers are eagerly exploring methods to enhance the geometric capabilities of MLLMs.

To enhance the geometric capabilities of MLLMs, an important step is to construct corresponding high-quality data~\cite{gao2023gllava,DBLP:conf/nips/ZhouLX0SMMEYYZG23,DBLP:conf/emnlp/ChenLQLLCL22}. 
Nevertheless, current data often suffer from two main issues.
On the one hand, most open-source datasets are quite challenging, making it difficult for models to directly learn geometric capabilities from them~\cite{DBLP:conf/icml/BengioLCW09,DBLP:conf/acl/XuZMWXZ20}. 
For instance, the UniGEO~\cite{DBLP:conf/emnlp/ChenLQLLCL22} dataset consists of problems extracted from high school textbooks, but the models have not been exposed to the corresponding foundational knowledge.
On the other hand, current data augmentation techniques~\cite{gao2023gllava}, using ChatGPT-3.5 to adjust numerical values in the text, fail to harmonize these changes with the corresponding values in images.
Consequently, mismatches between the altered text and images can bewilder the model and impede its learning process~\cite{DBLP:conf/emnlp/HesselHFBC21,DBLP:conf/iclr/YaoHHLNXLLJX22}.

In this paper, we address the aforementioned issues by introducing a straightforward and efficient pipeline for generating geometric problem data.
Our objectives are two-fold: (1) to create geometric problems that facilitate the model's acquisition of basic geometric concepts, and (2) to ensure that the image and the text of the generated geometric problems are well-aligned.
In detail, we first employ GPT-4V to create a collection of simplified geometric problems based on open-source datasets.
Subsequently, we harness the capabilities of GPT-4~\cite{openai2023gpt4} to generate $K$ individual pieces of Wolfram\footnote{The Wolfram is a computational language designed to handle various computing and data analysis tasks, possessing a formidable capability for geometric visualization.} code for each geometric problem previously crafted. 
The code is then executed to produce $K$ distinct geometric images.
Finally, GPT-4V is employed to score these images, allowing us to select the best one that optimally aligns with the associated textual descriptions.

Through the above pipeline, we generate a dataset comprising 4.9K geometric problems characterized by simplicity and image-text matching.
We then mix our generated problems with 19K problems from open-source datasets to formulate a dataset with uniform difficulty, named GeoGPT4V.
We have conducted comprehensive experiments on the geometry problem subset of MathVista~\cite{lu2024mathvista} and MathVision~\cite{wang2024measuring} datasets, two commonly used datasets for multi-modal math. 
Our experimental results show that models of various sizes and types can achieve significant improvements in geometric capabilities after training with our dataset (achieving 58.2\% and 33.8\% relative improvement for LLaVA-1.5-7B~\cite{liu2023improvedllava} and ShareGPT4V-7B~\cite{chen2023sharegpt4v}, respectively, on Geometry problem solving (GPS) minitest split of MathVista), which validates the effectiveness of our approach.

In conclusion, the contributions of this paper are summarized as follows:
\begin{itemize}[topsep=0pt,itemsep=0pt,parsep=0pt,leftmargin=*]
    \item We first introduce a novel pipeline capable of automatically generating simple geometric data with aligned image-text pairs.
    
    \item We have open-sourced the 4.9K dataset generated by our pipeline, along with the checkpoints of models trained on GeoGPT4V, to facilitate the community’s growth and development.
    
    \item Extensive experiments have consistently shown that GeoGPT4V effectively enhances the multi-modal geometric capabilities of models of various types and sizes.
    
\end{itemize}

\section{Related Work} 
In this section, we delve into related studies from two perspectives: multi-modal large language models and mathematical problem solving.

\paragraph{Multi-modal Large Language Models.}
With the rapid advancement of LLMs, the research community has started to develop multi-modal extensions of these models, known as MLLMs~\cite{Qwen-VL, openai2023gpt4v, liu2023llava}.
These MLLMs integrate visual information with linguistic data, enhancing their capabilities significantly~\cite{deepseekvl, monkey, mplug2,instructblip}.
Closed-source model, such as GPT-4V~\cite{openai2023gpt4v}, Gemini~\cite{geminiteam2023gemini}, and Qwen-VL-Max~\cite{Qwen-VL}, have demonstrated remarkable proficiency in image comprehension and cognitive tasks.
For open-source models, LLaVA~\cite{liu2023llava,liu2023improvedllava,liu2024llavanext} utilizes linear projection to bridge the visual encoder and the language model, achieving commendable performance in multi-modal tasks. 
Building upon the LLaVA architecture, ShareGPT4V~\cite{chen2023sharegpt4v} employs high-quality instructional data to further enhance model capabilities.
Moreover, InternVL-Chat~\cite{chen2023internvl} upscales its visual encoder to 6 billion parameters.
InternLM-XComposer2~\cite{internlmxcomposer2} excels in free-form text-image composition and understanding.
Although these MLLMs have shown powerful visual capabilities, MLLMs still confront challenges when it comes to mathematical problem-solving, as highlighted by recent studies~\cite{wang2024measuring, lu2024mathvista, mmmu}.
\begin{figure*}[htbp]
\centering
\includegraphics[width=1\textwidth]{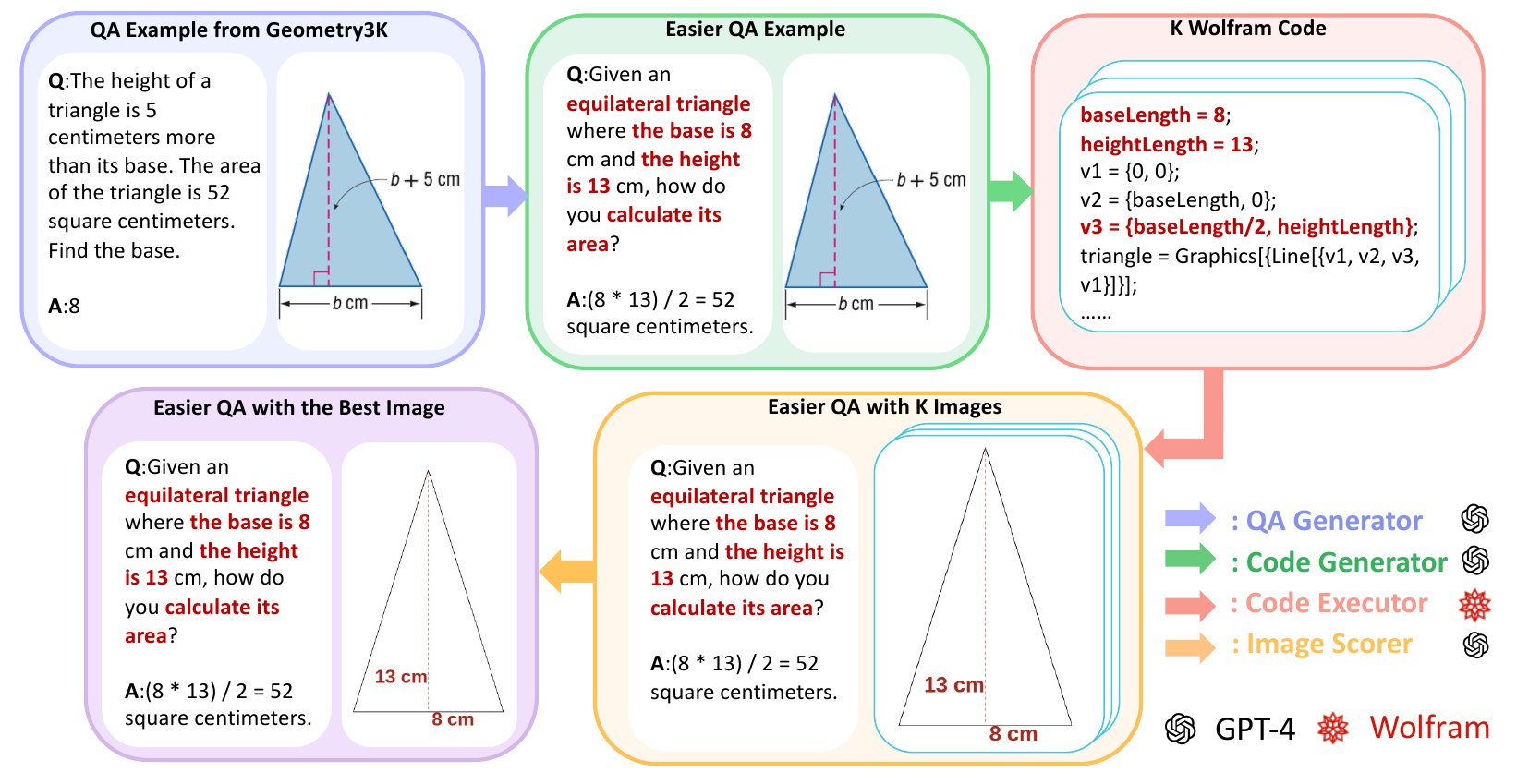}
\caption{\textbf{Pipeline of our geometric data generation.}
During the first step, we employ GPT-4V to generate simplified geometric question-answer pairs based on open-source datasets.
We \textcolor[HTML]{b02412}{\textbf{highlight}} the simplified parts compared to the original questions.
During the second step, we employ GPT-4 to generate $K$ Wolfram code for each question-answer pair.
During the third step, we execute $K$ code to obtain $K$ images.
During the fourth step, we employ GPT-4V to score the degree of alignment between the generated images and the questions.
We choose the image with the highest score.
Finally, we can obtain simplified and image-text matching geometric problems.
}
\label{fig:method_pipline}
\end{figure*}

\paragraph{Mathematical Problem Solving.}
The remarkable reasoning capabilities of LLMs have spurred researchers to harness them for solving mathematical problems~\cite{DBLP:journals/corr/abs-2308-07921,DBLP:journals/corr/abs-2402-03300, stepbystep, DBLP:conf/emnlp/ZhaoXKHX23}.
In the realm of pure text-based mathematical tasks, WizardMath~\cite{luo2023wizardmath} enhances model performance by refining instructions through a process of downward and upward instruction evolution. 
MetaMath~\cite{yu2023metamath} approaches the challenge by bootstrapping mathematical questions and rewriting them from various perspectives to improve understanding and problem-solving.
However, as previous studies have found, humans’ accuracy significantly decreases when solving geometry problems without images~\cite{dblp:conf/acl/chentqllxl21}. 
Therefore, geometry problems necessitate the visual perception abilities of multi-modal models to fully comprehend and solve them.
UniGeo~\cite{DBLP:conf/emnlp/ChenLQLLCL22} addresses this by compiling geometry problems from high school textbooks and introducing a unified multitask geometric transformer framework to tackle calculation and proving problems simultaneously in the form of sequence generation.
G-LLaVA~\cite{gao2023gllava} leverages ChatGPT-3.5 to create geometric question-answer pairs and to rewrite the textual content within questions. 
Nevertheless, this approach of textual rewriting alone may result in discrepancies between images and text, leading the model to produce incorrect or unrealistic outputs~\cite{DBLP:journals/corr/abs-2310-14566}.
This highlights the ongoing challenge of aligning textual and visual information in multi-modal mathematical problem-solving.

\section{Method}
\label{sec:method}

In this section, we will elaborate on the pipeline we have constructed. 
An overview of our pipeline is depicted in Figure~\ref{fig:method_pipline}. 
Specifically, our process includes: 
(1) generating new question-answer pairs (Section~\S\ref{sec:3.1}), 
(2) producing corresponding geometric images (Section~\S\ref{sec:3.2}), and 
(3) scoring and filtering based on the image-text matching degree (Section~\S\ref{sec:ranking}).

Formally, the original data from the open-source datasets can be represented as $D = \{Q, A, I\}$, where $Q$ represents the question, $A$ represents the answer, and $I$ represents the image.

\subsection{Question-Answer Pairs Generation}
\label{sec:3.1}

Due to the prevalence of more challenging geometric problems in open-source datasets, to facilitate our model’s learning of basic geometric concepts, we initially simplify these difficult problems to generate easier geometric question-answer (QA) pairs.

In detail, we utilize GPT-4V~\cite{openai2023gpt4v} to generate QA pairs from the dataset $D = \{Q, A, I\}$.
We instruct GPT-4V to craft simplified problems that are derived from the original geometric QA pairs to acquire QA pairs containing fundamental geometric concepts.
In detail, we prompt GPT-4V to consider these three perspectives: (1) generating lead-up problems, (2) generating sub-problems, and (3) incorporating the conclusions from the answer into the conditions of the question, which can reduce the complexity of the question.
To prevent GPT-4V from generating the same simplified questions, we also ask GPT-4V to generate questions that are as diverse as possible.
Additionally, for efficiency, the instruction also asks GPT-4V to generate textual descriptions of images aimed at supporting the subsequent phase of image generation.
The detailed prompt can be found in Appendix~\ref{sec:qa_prompt}.

In practice, we generate $N$ ($N = 3$) new data points based on a single original data point to improve efficiency and reduce API costs.
After this phase, the data we obtain can be formally represented as $\tilde D_1 = \{\tilde Q, \tilde A, \tilde {Des}\}$ where $\tilde {Des}$ represents the image description.

\subsection{Geometric Images Generation}
\label{sec:3.2}

It is important to highlight that the newly generated QA pairs may not correspond directly to the original images, which could hurt the model’s learning process. 
To ensure congruity between the textual content and the visual aspects, it is essential to produce new images that align with the generated QA pairs. 
To address this issue, we employ Wolfram, a powerful software tool capable of executing code to generate geometric images.

In detail, we utilize GPT-4~\cite{openai2023gpt4} to generate Wolfram code based on the dataset $\tilde D_1$.
Firstly, we feed the questions, answers, and image descriptions as prompts to GPT-4 to generate Wolfram code.
During the generation process, we instruct GPT-4 to explicitly name all variables within the code, with the aim of facilitating a clearer understanding and assisting GPT-4 in recognizing the relationships between code elements and the given questions.
The detailed prompt can be found in Appendix~\ref{sec:code_prompt}.
Finally, we execute the Wolfram code, resulting in the generation of new images.

In practice, it is noticed that employing GPT-4 to generate code is unstable.
Thus, we generate $K$ ($K = 3$) distinct code from the same data to increase the probability of obtaining the correct code.
Consequently, we can obtain $K$ distinct images corresponding to $K$ code.
It can be represented as $\tilde D_2 = \{\tilde Q, \tilde A, \tilde I^{(1)},\tilde I^{(2)}, \dots,  \tilde I^{(K)}\}$, where $\tilde I^{(i)}$ represents the $i$-th image generated for each question.

\subsection{Scoring and Filtering}
\label{sec:ranking}

After generating $K$ images using Wolfram for each question, we need to select the most suitable one to be used as the final image in our dataset.

Concretely, we employ GPT-4V to assign a score ranging from 0 to 1 that reflects the degree of correspondence between an image generated for the question and the question itself; a higher score signifies a stronger alignment. 
To augment the scoring proficiency of GPT-4V, drawing inspiration from the Chain-of-Thought~\cite{DBLP:conf/nips/Wei0SBIXCLZ22}  , we instruct GPT-4V to articulate the rationale underlying its evaluation before determining the ultimate score.
The detailed prompt can be found in Appendix~\ref{sec:scoring_prompt}.

Finally, for each question associated with $K$ distinct generated images, we obtain $K$ corresponding scores. 
For each question, we retain the image with the highest score as $\tilde I$. 
Note that, if this score is less than 0.9, we consider that the image for this question has not been well-generated, and we discard the question.
Consequently, we compile a dataset $\tilde D = \{\tilde Q,\tilde A, \tilde I\}$ that consists of questions that are simpler and exhibit a stronger alignment between the images and the associated text.

\section{Data Analysis}
\begin{table}[htbp]
\centering
 \renewcommand\arraystretch{0.8} 
\begin{tabular}{lr}
\toprule
\multicolumn{1}{l}{Datasets}                               & Samples              \\ 
\toprule
\multicolumn{2}{l}{\textit{\textbf{Open-source Datasets}}} \\ 
\midrule
\multicolumn{1}{l}{ChartQA}                               & 7398                 \\ 
\multicolumn{1}{l}{UniGEO-Calculation}                     & 3499                 \\ 
\multicolumn{1}{l}{Geometry3K}                             & 2101                 \\ 
\multicolumn{1}{l}{GeoQA+}                                 & 6026                 \\ 
\midrule
\multicolumn{2}{l}{\textit{\textbf{Generated Datasets}}}            \\ 
\midrule
\multicolumn{1}{l}{UniGEO-Proving\_Enhanced}               & 1810                 \\ 
\multicolumn{1}{l}{Geometry3K\_Enhanced}                   & 1909                 \\ 
\multicolumn{1}{l}{GeoQA\_Enhanced}                       & 1212                 \\ 
\bottomrule
\end{tabular}
\caption{\textbf{
The datasets used in the GeoGPT4V dataset.}
Column “Samples” is the number of image-text pairs in each dataset.
It is worth noting that we only use the training sets of open-source datasets.
}
\label{tab:training_datatsets}
\end{table}
In this section, we will present a comprehensive statistical analysis (Section~\S\ref{sec:4.1}) and evaluation (Section~\S\ref{sec:4.2}~\S\ref{sec:4.3}) of the datasets generated through our pipeline.

\subsection{Datasets}
\label{sec:4.1}
In this study, to minimize costs, we selected the first 1,500 samples from the training sets of the UniGEO-Proving~\cite{DBLP:conf/emnlp/ChenLQLLCL22}, Geometry3K~\cite{dblp:conf/acl/lugjqhlz20}, and GeoQA~\cite{dblp:conf/acl/chentqllxl21} to create UniGEO-Proving\_Enhanced, Geometry3K\_Enhanced, and GeoQA\_Enhanced for validating the effectiveness of our method.
Subsequently, we combine the generated geometric problems with those from open-source datasets, including ChartQA~\cite{dblp:conf/acl/masryltjh22}, UniGEO-Calculation~\cite{DBLP:conf/emnlp/ChenLQLLCL22}, the original Geometry3K~\cite{dblp:conf/acl/lugjqhlz20}, and GeoQA+~\cite{dblp:conf/coling/caox22}, to form a new dataset with uniform difficulty levels, dubbed GeoGPT4V.
A detailed breakdown of the datasets is provided in Table~\ref{tab:training_datatsets}.

\subsection{Difficulty Evaluation}
\label{sec:4.2}
As mentioned in Section~\S\ref{sec:method}, our pipeline will take original data $D$ as input and output generated data $\tilde D$.
We aim to generate easier data than the original one to facilitate model learning of basic geometric knowledge.
This section demonstrates the efficacy of our pipeline by comparing the difficulty levels of $D$ and $\tilde D$.

We initiate this by forming a data pair $P_1 = \{D, \tilde D\}$ and utilize GPT-4V to assess the relative difficulty of the data points. 
To mitigate the bias that GPT-4V may have due to the presentation order, we also consider the pair $P_2= \{\tilde D, D\}$, obtained by swapping the order of the data points. 
If GPT-4V produces different outputs based on $P_1$ and $P_2$, we conclude that the difficulty of $D$ and $\tilde D$ is equal.
A detailed prompt can be found in Appendix~\ref{sec:difficulty_prompt}.

In practice, we randomly sample 500 pairs of generated and corresponding original data points. The outcome, presented in Figure~\ref{fig:data_difficulty}, reveals that over 80\% of the questions in the generated dataset are of equal or lesser difficulty compared to the original questions. This indicates that our pipeline is successful in generating data that is simpler than the original dataset.

\subsection{Image-text Matching Evaluation}
\label{sec:4.3}
As mentioned in the previous section, the alignment between text and images is a critical aspect of geometric problem data.
To illustrate that the generated images are better suited for the simplified problems than the original images, we replace the generated images with the original image for each question, resulting in new data $\tilde D ^ {\prime} = \{\tilde Q, \tilde A, I\}$.
Consequently, in this section, we will compare the level of image-text matching in our generated data $\tilde D$ with $\tilde D ^ {\prime}$ and the QA data produced by prior methods -- G-LLaVA~\cite{gao2023gllava}.
Similar to the score function in Section~\S\ref{sec:ranking}, we employ GPT4-V to score the degree of alignment between the images and the questions.

\begin{figure}[h]
    \centering
    \subfloat[]{\includegraphics[width=0.33\linewidth]{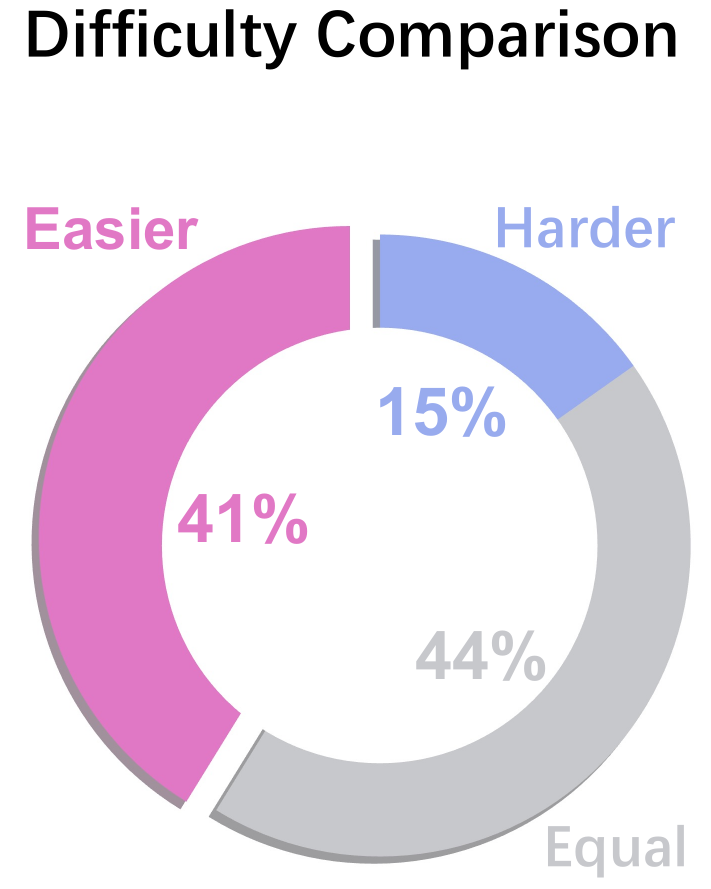}\label{fig:data_difficulty}}\hfill
    \subfloat[]{\includegraphics[width=0.65\linewidth]{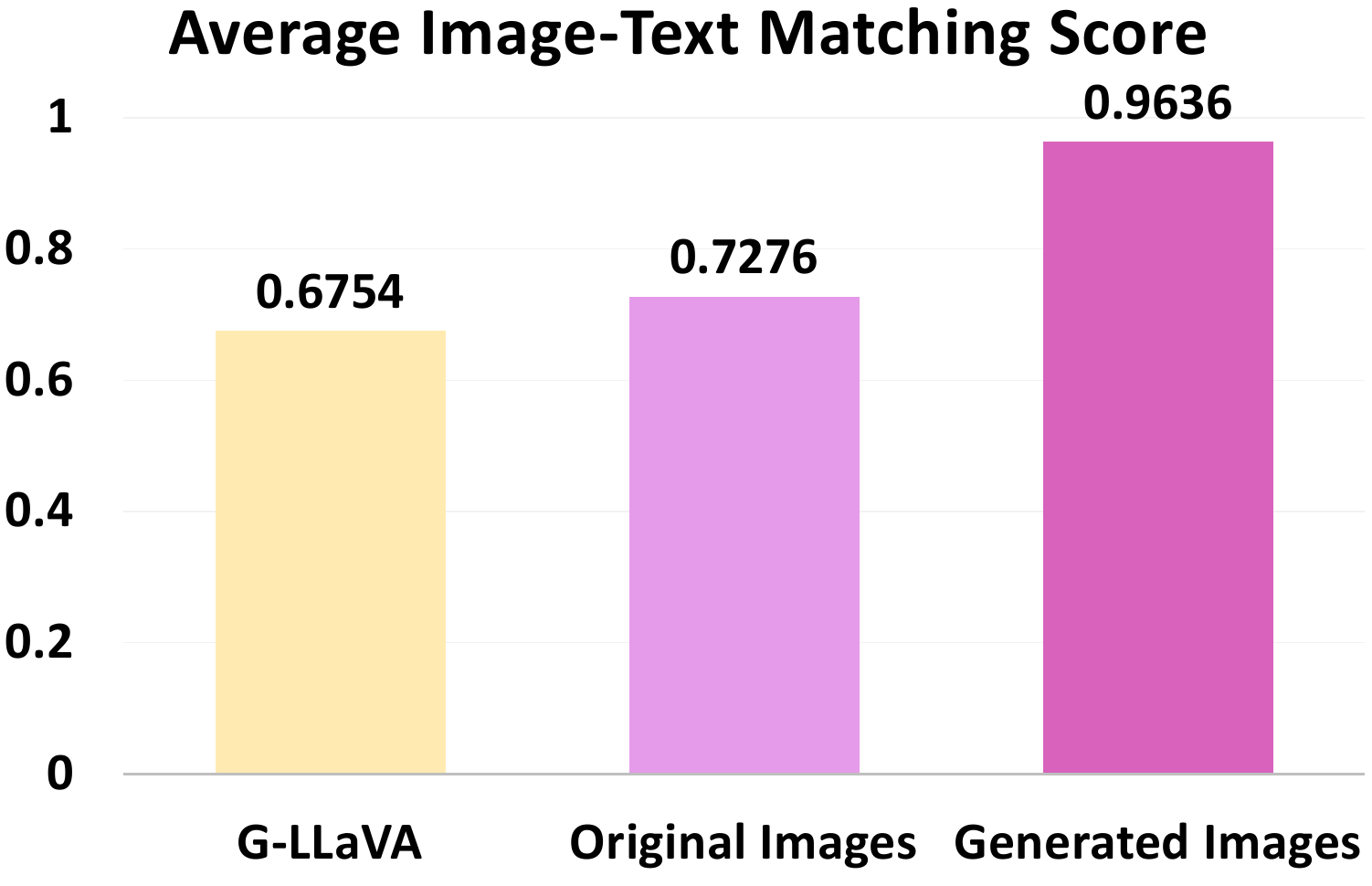}\label{fig:data_matching}}
    \caption{\textbf{The data analysis results.}
    This chart illustrates the simplicity and image-text matching attributes of our dataset.
    Figure (a) is a comparison chart of the difficulty between the generated and original data.
    In this figure, ``Easier'' represents that the generated data is easier than the original data;
``Harder'' represents that the generated data is harder than the original data;
``Equal'' represents that the generated and original data have the same difficulty level.
    Figure (b) shows the average image-text matching scores for the three data types.
    ``Generated Images'' represents our generated data.
    ``Original Images'' represents the data obtained by replacing generated images in generated data with original images.
    }
    \label{fig:data_ann}
\end{figure}

In detail, we randomly select 500 data points for each dataset and show the average scores of the three datasets in Figure~\ref{fig:data_matching}.
The results indicate that our generated data, $\tilde D$, exhibits a significantly higher degree of image-text matching than $\tilde D^ {\prime}$, as well as the dataset enhanced by G-LlaVA (0.9636 for $\tilde D$, 0.7276 for $\tilde D^ {\prime}$, and 0.6754 for G-LlaVA).
Moreover, it is observed that G-LlaVA’s image-text matching score is the lowest, which confirms our hypothesis that simply scaling the size of numbers within problems is an inappropriate approach.
\section{Experiment}
\begin{table*}[]

\centering
\small
 \renewcommand\tabcolsep{2.4pt} 
 \renewcommand\arraystretch{1.1} 
\begin{tabular}{lr|ccc|cccccccccc}
\toprule
\multicolumn{1}{l}{\multirow{2}{*}{Model}} & \multicolumn{1}{c|}{\multirow{2}{*}{Size}} & \multicolumn{3}{c|}{MathVista}                                 & \multicolumn{10}{c}{MathVision}                                                                                                                                                                                                                                            \\ \cline{3-15} 
\multicolumn{1}{c}{}                       & \multicolumn{1}{c|}{}                        & GPS   & GEO & AVG     & AnaG  & CombG & DescG & GrphT & Angle & Area  & Len   & SolG  & TransG& AVG \\ 
\toprule
LLaVA-1.5                                    & 7B                                          & 20.67$^{*}$ & 20.92$^{*}$&20.80$^{*}$  & 7.1   & 7.1   & 7.7   & 10    & 15.6  & 10.2  & 9.8   & 5.3   & 4.8 &8.62   \\ 

LLaVA-1.5                                    & 13B                                        & 24.04$^{*}$ & 23.85$^{*}$&23.95$^{*}$  & 14.3  & 9.1   & 13.5  & 5.6   & 10.4  & 12.6  & 14.7  & 11.5  & 10.7&11.38   \\ 
\midrule
LLaVA-1.5-G                                & 7B                                          & \color{up}32.69 & \color{up}32.22&\color{up}32.46  & \color{up}9.52  & \color{up}16.88 & \color{up}9.62  & \color{up}21.11 & \color{up}19.08 & \color{up}11.06 & \color{up}17.15 & \color{up}9.43  & \color{up}15.48&\color{up}14.37  \\ 
LLaVA-1.5-G                                & 13B                                         & \color{up}36.54 & \color{up}37.24 &\color{up}36.89 & \color{up}15.48  & \color{up}14.29 & \color{down}12.50  & \color{up}18.89 & \color{up}19.65 & \color{up}13.60   & \color{up}18.49   & \color{down}9.02  & \color{up}11.31&\color{up}15.14   \\ 
\midrule
ShareGPT4V                                   & 7B                                        & 21.63$^{*}$ & 20.50$^{*}$&21.07$^{*}$  & 3.6   & 10.1  & 11.5  & 14.4  & 16.2  & 11.8  & 12.3  & 9.8   & 11.3 &11.22  \\ 
ShareGPT4V                                   & 13B                                        & 27.4$^{*}$  & 27.62$^{*}$&27.51$^{*}$ & 15.5  & 10.7  & 11.5  & 8.9   & 11.6  & 13    & 17.4  & 10.3  & 12.5  &12.38 \\ 
\midrule
ShareGPT4V-G                               & 7B                                        & \color{up}32.69 & \color{up}31.80&\color{up}32.25   & \color{up}11.90  & \color{up}12.99 & \color{down}9.62  & \color{up}16.67 & \color{up}17.34 & \color{up}13.60  & \color{up}17.59 & \color{up}10.25 & \color{up}11.31 &\color{up}13.47 \\ 
ShareGPT4V-G                               & 13B                                       & \color{up}43.27 & \color{up}42.26&\color{up}42.77  & \color{up}22.62 & \color{down}9.74  & \color{up}13.46 & \color{up}11.11 & \color{up}19.08 & \color{up}15.80  & \color{down}13.81 & \color{down}9.02  & \color{up}13.69&\color{up}14.26  \\ 
\midrule
InternVL\dag                                   & 40B                                         & 61.1  & 61.1&61.10 & 16.67$^{*}$  & 12.99$^{*}$  & 15.38$^{*}$  & 13.33$^{*}$   & 4.62$^{*}$  & 5.60$^{*}$    & 6.46$^{*}$  & 9.84$^{*}$  & 10.71$^{*}$ &10.62$^{*}$  \\ 
InternVL-G\dag                                   & 40B                                        & \color{up}64.42  & \color{up}63.60&\color{up}64.01  & 16.67  & \color{up}18.18  & \color{down}13.46  & \color{up}16.67   & \color{up}23.12  & \color{up}18.40    & \color{up}18.93  & \color{up}11.89  & \color{up}23.21 &\color{up}17.84  \\ 
\midrule
\multicolumn{15}{c}{Closed-source Models}\\ 
\midrule
Qwen-VL-Plus & -&38.5&39.3&38.90&17.9&12.7&15.4&8.9&11.6&6.4&10.0&14.3&11.31&12.06 \\
Qwen-VL-Max & -&-&-&-&19.1&16.9&16.4&12.2&13.3&14.2&19.8&11.5&17.3&15.61 \\
Gemini-1.0-Pro & -&40.4&41.0&40.70& 10.7&20.1&20.2&21.1&19.1&19.0&20.0&14.3&20.8&18.37\\
Gemini-1.0-Ultra & -&56.2&55.6&55.90&-&-&-&-&-&-&-&-&-&- \\
GPT-4V & -&50.5&51.0&50.75&32.1&21.1&22.1&14.4&22.0&22.2&20.9&23.8&25.6&22.69 \\ 
\bottomrule
\end{tabular}
\caption{\textbf{Overall results of different models on the MathVista and MathVision.}
We present the detailed scores for all the tasks related to geometry such as ``GPS'' and ``AnaG'', as well as the average score over these tasks in two benchmarks denoted as ``AVG''. Due to limited space, we utilize abbreviations for these geometry-related tasks and illustrate the detailed task name in the Appendix \ref{sec:Detailed_TI}.
For the model trained with GeoGPT4V, score increases are marked in \textcolor{up}{red} 
compared to the original model.
$^{*}$ indicates our re-implemented test results missed in benchmarks or origin papers.
InternVL\dag represents the abbreviation for InternVL-Chat-V1.2-Plus.
The suffix “-G” to the model name indicates a model trained on the GeoGPT4V.
For better comparison, we also demonstrate results for five representative closed-source MLLM models.}
\label{tab:main_models}
\end{table*}

In this section, we conduct experiments to answer the following research questions (RQ):
\begin{itemize}[leftmargin=*]
    \item \textbf{RQ1}: Can GeoGPT4V dataset improve geometric capabilities of different models?
    \item \textbf{RQ2}: Are the generated images better than the original images for model learning?
    \item \textbf{RQ3}: Is it necessary to score and filter the generated images?
    \item \textbf{RQ4}: Is the improvement solely due to the original dataset?
\end{itemize}

\subsection{Experimental Setup}

\paragraph{Benchmarks.}
We utilize two widely used benchmarks, which encompass numerous multi-model geometric problems, to evaluate the effectiveness of our proposed GeoGPT4V dataset. The detailed information of these benchmarks is as follows:
\begin{itemize}[leftmargin=*]
    \item \textit{MathVista~\cite{lu2024mathvista}} is a mathematical reasoning benchmark in visual contexts.
    It includes diverse visual contexts, such as natural images, geometry diagrams, charts, etc.
    MathVista includes multiple-choice questions as well as open-ended questions.
    The MathVista test set comprises 5141 examples without ground truth answers and 
    provides 1000 examples with ground truth answers known as MathVista testmini.
    
    \item \textit{MathVision~\cite{wang2024measuring}} is a more challenging multi-modal mathematical benchmark than MathVista.
    It categorizes all mathematical problems into five difficulty levels and 16 distinct tasks.
    MathVision also consists of multiple-choice questions and open-ended questions.
    The MathVision test set comprises 3040 examples with ground truth answers.
\end{itemize}

\paragraph{Evaluation Method.}
We strictly follow the evaluation method proposed in MathVista~\cite{lu2024mathvista} and MathVision~\cite{wang2024measuring}. Firstly, we utilize ChatGPT-3.5 to extract the ultimate response from model outputs in MathVista, while employing regular expressions with MathVision for the same purpose. Consequently, we report the accuracy of the answers as the score for performance evaluation.

\paragraph{Baseline Models.}
We train the following main-stream open-source models using our proposed GeoGPT4V dataset, with model sizes including 7B, 13B, and 40B.

\begin{itemize}[leftmargin=*]

    \item \textit{LLaVA-1.5~\cite{liu2023llava, liu2023improvedllava}} utilizes linear layers to connect the vision encoder and the large language model (LLM).
    In the pre-training stage, LLaVA-1.5 keeps the vision encoder and the LLM frozen, and only trains linear layers.
    In the fine-tuning stage, it freezes the vision encoder and trains the linear layers and the LLM.

    \item \textit{ShareGPT4V~\cite{chen2023sharegpt4v}} has an architecture similar to LLaVA's. 
    However, in the pre-training stage of ShareGPT4V, both the vision encoder and the language model remain unfrozen. 
    The training data is high-quality, detailed description data generated by GPT-4V.
    
    \item \textit{InternVL-Chat-V1.2-Plus~\cite{chen2023internvl}} utilizes the InternViT~\cite{chen2023internvl} as its visual encoder, which has 6 billion parameters.
    What's more, it scales LLM to 34B and utilizes a fine-tuning dataset with 12 million samples.

\end{itemize}

\begin{table*}[]

\centering
\small
 \renewcommand\tabcolsep{3pt} 
 \renewcommand\arraystretch{1.2} 
\begin{tabular}{l|ccc|cccccccccc}
\toprule
\multicolumn{1}{l|}{\multirow{2}{*}{Model}} & \multicolumn{3}{c|}{MathVista}                                 & \multicolumn{10}{c}{MathVision}                                                                                                                                                                                                                                            \\ \cline{2-14} 
 \multicolumn{1}{c|}{}                        & GPS   & GEO& AVG      & AnaG  & CombG& DescG & GrphT & Angle & Area  & Len   & SolG  & TransG& AVG \\ 
\toprule
LLaVA-1.5-7B                                          & 20.67$^{*}$ & 20.92$^{*}$&20.80$^{*}$   & 7.1   & 7.1   & 7.7   & 10    & 15.6  & 10.2  & 9.8  & 5.3   & 4.8 &8.62   \\ 
\midrule
- Image Generation                                        & 30.77 & 30.96 &30.87& 8.33  & 14.94 & 8.65  & 15.56 & 17.34 & \textbf{12.20} & 14.48 & 7.79  & \textbf{19.05} &13.15 \\
- Image Scoring                                         & \textbf{33.65} & 31.80 &\textbf{32.73} & \textbf{9.52}  & 15.48 & 9.62  & 20.00 & 17.34 & \textbf{12.20} & 15.59 & 6.56  & 15.48 &13.54 \\
\midrule
GeoGPT4V                                          & 32.69 & \textbf{32.22}&32.46& \textbf{9.52}  & \textbf{16.88} & \textbf{9.62}  & \textbf{21.11} & \textbf{19.08} & 11.06 & \textbf{17.15} & \textbf{9.43}  & 15.48  &\textbf{14.37}   \\ 
\bottomrule
\end{tabular}
\caption{\textbf{Ablation for image generation and image scoring.}
``- Image Generation'' denotes the exclusion of newly generated geometric images. ``- Image Scoring'' signifies the random selection of generated images, rather than utilizing GPT4V to score and choose them.
For comparison, we also represent the results from the official LLaVA-1.5-7B model in the first line and GeoGPT4V in the last line.
\textbf{Bold results} indicate the best results for all models.
$^{*}$ indicates our re-implemented test results missed in benchmarks or origin papers.
}
\label{tab:ablation_origin_images}
\end{table*}

\paragraph{Implementation Details.}
For data generation, we employ  ``gpt-4-vision-preview'' and ``gpt-4-1106-preview'' API provided by OpenAI for GPT-4V and GPT-4.
For model training, all the models are trained on NVIDIA A100 GPUs with PyTorch version 2.0.1.
To ensure the fair comparison, we keep the training parameters consistent with those specified by the model's original authors and train the models for one epoch.
Detail training parameters are demonstrated in Appendix~\ref{sec:training_parameters}.

\subsection{Main Results (RQ1)}

We evaluate the performance of various open-source models on MathVista testmini (short as MathVista) and MathVision test (short as MathVision) benchmarks after training on the GeoGPT4V dataset to demonstrate our proposed method's effectiveness.
For convenience, we append the suffix ``-G'' to the model name to indicate a model trained on the GeoGPT4V dataset, such as ``LLaVA-1.5-G''.
Since our method focuses on geometric data, we present detailed scores for all the tasks related to geometry and the average score over these tasks in Table~\ref{tab:main_models}.
The complete set of scores can be found in Appendix~\ref{sec:detail_mathvista} and ~\ref{sec:detail_mathvision}.
In Appendix~\ref{sec:comparsion_other_models}, we compare the geometric capabilities of our best model, InternVL-Chat-V1.2-Plus-GeoGPT4V, with other open-source and closed-source models.

The experimental results from Table~\ref{tab:main_models} indicate that our dataset can effectively improve different models' geometric capabilities.
First of all, our proposed GeoGPT4V has exhibited an improvement in the average scores across all geometry-related tasks on both MathVista and MathVision benchmarks, indicating that GeoGPT4V can enhance the model's general geometry performance.
Moreover, our proposed GeoGPT4V has brought improvements to most geometry-related tasks in both benchmarks
in all scales and types of models.
Furthermore, our GeoGPT4V significantly bridges the gap in geometric capabilities between open-source and closed-source models, except InternVL-Chat-V1.2-Plus, which has already employed a substantial amount of customized fine-tuning datasets.

\subsection{In-depth Analysis}
\begin{table*}[]
\centering
\renewcommand\arraystretch{0.9} 
\scalebox{0.95}{
\begin{tabular}{l|lll}
\toprule
Name & Base & Replace & Mix \\ 
\toprule
\multirow{5}{*}{Datasets} & ChartQA & ChartQA & ChartQA \\ 
 & UniGeo-Calculation & UniGeo-Calculation & UniGeo-Calculation \\ 
 & Geometry3K & Geometry3K\_Replace & Geometry3K\_Mix \\ 
 & GeoQA+ & GeoQA+\_Replace & GeoQA+\_Mix \\  
 & UniGeo-Proving & UniGeo-Proving\_Replace & UniGeo-Proving\_Mix \\ 
 \bottomrule
\end{tabular}}
\caption{\textbf{Dataset settings for experiments comparing open-source data and generated data.} 
The suffix ``Replace'' indicates that we replace the corresponding original data with generated data.
The suffix ``Mix'' indicates that we mix the original data with generated data.}

\label{tab:ablation_datasets}
\end{table*}

\begin{table*}[]

\centering

\scalebox{0.85}{
 \renewcommand\tabcolsep{3pt} 
 \renewcommand\arraystretch{1} 
\begin{tabular}{l|ccc|cccccccccc}
\toprule
\multicolumn{1}{c|}{\multirow{2}{*}{Datasets}} & \multicolumn{3}{c|}{MathVista}                                 & \multicolumn{10}{c}{MathVision}                                                                                                                                                                                                                                            \\ \cline{2-14} 
\multicolumn{1}{c|}{}                        & GPS   & GEO &AVG     & AnaG  & CombG& DescG & GrphT & Angle & Area  & Len   & SolG  & TransG &AVG\\ 
\toprule
 Base                                           & 29.33 & 28.03&28.68  & 10.71  & \textbf{15.91} & 8.65  & 12.22 & 16.67 & 11.80 & 13.59 & 8.20  & 16.07&12.65  \\ 
Replace                                          & 33.17 & 32.64 &32.91 & 7.14  & 14.94 & 6.73  & \textbf{20.00} & \textbf{20.81} & 10.80 & \textbf{15.14} & \textbf{10.25}  & 14.29  &13.34\\ 
Mix                                         & \textbf{33.52} & \textbf{34.31}&\textbf{33.92} & \textbf{11.90}  & 15.58 & \textbf{10.58}  & 14.44 & 17.34 & \textbf{12.40} & 14.48 & 9.43  & \textbf{16.07}  &\textbf{13.58}   \\ 
\bottomrule
\end{tabular}}
\caption{\textbf{Comparison of the effects with and without using the generated datasets.}
\textbf{Bold results} indicate the best results for all models.
}
\label{tab:ablation_open_source}
\end{table*}
To comprehensively analyze the effectiveness of GeoGPT4V, we design a series of analyzing experiments from various perspectives.
Firstly, we design ablation experiments from the standpoint of the efficacy of generating new geometric images and selecting generated images with GPT4V scores.
Subsequently, we conduct experiments to demonstrate the substantial performance improvement brought by GeoGPT4V stemming from the generated data rather than the utilization of open-source data. 
Due to resource and space limitations, we leverage LLaVA-1.5-7B for analytical experiments and conduct evaluations on both MathVista and MathVision.

\subsubsection{Effect of Generating New Images (RQ2)}
We validate the effectiveness of the newly generated geometric images by replacing the images generated in GeoGPT4V with their original counterparts and training the model on them.
In detail, we firstly substitute the newly generated images from GeoGPT4V with the original images while retaining the simplified questions generated, formulating a new dataset denoted as $\tilde D^ {\prime}$.
Subsequently, we train the LLaVA-1.5-7B model on $\tilde D^ {\prime}$ and compare its geometric capabilities with the model trained on GeoGPT4V.

Based on results demonstrated in Table~\ref{tab:ablation_origin_images}, we have following observations:  
Firstly, the model trained on $\tilde D^ {\prime}$ exhibits inferior performance compared to the model trained on GeoGPT4V, indicating the effectiveness of the newly generated images.
Secondly, the model trained on $\tilde D^ {\prime}$ demonstrates stronger performance than the model trained without the use of $\tilde D^ {\prime}$, thereby validating the efficacy of the easier QA pairs generated by our pipeline.

\subsubsection{Is Scoring Necessary? (RQ3)}

As mentioned in Section~\S\ref{sec:ranking}, $K$ images are scored, and the one with the highest score is selected from this set.
To demonstrate the necessity of scoring, 
we formulate a new dataset $\tilde D^ {\prime \prime}$ by directly modifying the selection method to randomly choose from the $K$ images while keeping all other aspects unchanged.
Consequently, we analyze the performance of the LLaVA-1.5-7B trained on $\tilde D^ {\prime \prime}$.

According to results demonstrated in Table~\ref{tab:ablation_origin_images}, we can find that the model trained on $\tilde D^ {\prime \prime}$ exhibits inferior performance compared to the model trained on GeoGPT4V.
The results indicate that the quality of the images obtained via ranking surpasses those chosen randomly.

\subsubsection{Are the Open-source Datasets Enough? (RQ4)}

To demonstrate performance improvements brought by GeoGPT4V  are not solely reliant on open-source data, we compare the performance of models trained using various combinations of open-source and our generated data.
In detail, as illustrated in Table~\ref{tab:ablation_datasets},
we construct three tiers of datasets. Firstly, we combine all open-source datasets to create the ``Base'' dataset. Subsequently, we replace the original data from the ``Base'' dataset with the data generated by our pipeline, resulting in the ``Replace'' dataset. Lastly, we mix the generated data with all the data from the ``Base'' dataset to form the ``Mix'' dataset.
It is notable that GeoQA is a subset of GeoQA+.
Thus we only use GeoQA+ in these three dataset settings, rather than using both GeoQA+ and GeoQA.

We finetune LLaVA-1.5-7B separately on these three datasets and evaluate their performance in Table~\ref{tab:ablation_open_source}, with observations as follows: Although the ``Base'' dataset, constructed using open-source data, provides moderate geometric capabilities, our ``Replace'' and ``Mix'' datasets exhibit even greater enhancements in geometric performance. This not only demonstrates the effectiveness of the data generated by our pipeline but also indicates that the improvements afforded by GeoGPT4V are not solely derived from open-source data.

\section{Conclusion} 

In this study, we propose a novel pipeline to enhance the geometric capabilities of MLLMs.
We have proposed data generation methods for multimodal geometric tasks involving problem simplification and the generation of images that match newly generated text.
Specifically, we use GPT4V and GPT4 to generate sub-problems or lead-up problems for given geometric tasks, along with the corresponding Wolfram code that can be executed to generate geometric images.
Based on the pipeline, we have generated 4.9K simplified and image-text matching geometric problems.
We mix our generated data with 19K open-source data to formulate a dataset with uniform difficulty, named GeoGPT4V.
After training on the GeoGPT4V dataset, various models have improved geometric scores on both MathVista and MathVision benchmarks.
The extensive experimental results demonstrate the effectiveness of the GeoGPT4V dataset.
We have open-sourced the GeoGPT4V dataset and the checkpoints of models trained on the GeoGPT4V dataset, with the aim of fostering the community's growth.

\section*{Limitations}

This paper focuses on the generation of geometric images.
We employ GPT-4 to generate Wolfram code, which can be executed to generate images.
However, this approach is unstable and may result in poor image quality.
That's why we use GPT-4V to score the images, which leads to more API calls and increased costs.

What's more, this paper only considers simplifying open-source geometric problems.
However, generating more complex problems is also worth considering, as it will generate more complex geometric images and help models improve complex reasoning capabilities.
Our future work will explore the more accurate generation of complex geometric images.

Finally, multi-modal mathematics is not limited to geometric problems.
It also includes tasks such as chart question answering and function question answering.
Generating richer charts and function images is also part of our future exploration work.

\bibliography{custom}

\begin{thebibliography}{40}
\providecommand{\natexlab}[1]{#1}

\bibitem[{Anil et~al.(2023)Anil, Borgeaud, Wu, Alayrac, Yu, Soricut, Schalkwyk, Dai, Hauth, Millican, Silver, Petrov, Johnson, Antonoglou, Schrittwieser, Glaese, Chen, Pitler, Lillicrap, Lazaridou, Firat, Molloy, Isard, Barham, Hennigan, Lee, Viola, Reynolds, Xu, Doherty, Collins, Meyer, Rutherford, Moreira, Ayoub, Goel, Tucker, Piqueras, Krikun, Barr, Savinov, Danihelka, Roelofs, White, Andreassen, von Glehn, Yagati, Kazemi, Gonzalez, Khalman, Sygnowski, and et~al.}]{geminiteam2023gemini}
Rohan Anil, Sebastian Borgeaud, Yonghui Wu, Jean{-}Baptiste Alayrac, Jiahui Yu, Radu Soricut, Johan Schalkwyk, Andrew~M. Dai, Anja Hauth, Katie Millican, David Silver, Slav Petrov, Melvin Johnson, Ioannis Antonoglou, Julian Schrittwieser, Amelia Glaese, Jilin Chen, Emily Pitler, Timothy~P. Lillicrap, Angeliki Lazaridou, Orhan Firat, James Molloy, Michael Isard, Paul~Ronald Barham, Tom Hennigan, Benjamin Lee, Fabio Viola, Malcolm Reynolds, Yuanzhong Xu, Ryan Doherty, Eli Collins, Clemens Meyer, Eliza Rutherford, Erica Moreira, Kareem Ayoub, Megha Goel, George Tucker, Enrique Piqueras, Maxim Krikun, Iain Barr, Nikolay Savinov, Ivo Danihelka, Becca Roelofs, Ana{\"{\i}}s White, Anders Andreassen, Tamara von Glehn, Lakshman Yagati, Mehran Kazemi, Lucas Gonzalez, Misha Khalman, Jakub Sygnowski, and et~al. 2023.
\newblock \href {https://doi.org/10.48550/ARXIV.2312.11805} {Gemini: {A} family of highly capable multimodal models}.
\newblock \emph{CoRR}, abs/2312.11805.

\bibitem[{Bai et~al.(2023)Bai, Bai, Yang, Wang, Tan, Wang, Lin, Zhou, and Zhou}]{Qwen-VL}
Jinze Bai, Shuai Bai, Shusheng Yang, Shijie Wang, Sinan Tan, Peng Wang, Junyang Lin, Chang Zhou, and Jingren Zhou. 2023.
\newblock Qwen-vl: A versatile vision-language model for understanding, localization, text reading, and beyond.
\newblock \emph{arXiv preprint arXiv:2308.12966}.

\bibitem[{Bengio et~al.(2009)Bengio, Louradour, Collobert, and Weston}]{DBLP:conf/icml/BengioLCW09}
Yoshua Bengio, J{\'{e}}r{\^{o}}me Louradour, Ronan Collobert, and Jason Weston. 2009.
\newblock \href {https://doi.org/10.1145/1553374.1553380} {Curriculum learning}.
\newblock In \emph{Proceedings of the 26th Annual International Conference on Machine Learning, {ICML} 2009, Montreal, Quebec, Canada, June 14-18, 2009}, volume 382 of \emph{{ACM} International Conference Proceeding Series}, pages 41--48. {ACM}.

\bibitem[{Cao and Xiao(2022)}]{dblp:conf/coling/caox22}
Jie Cao and Jing Xiao. 2022.
\newblock \href {https://aclanthology.org/2022.coling-1.130} {An augmented benchmark dataset for geometric question answering through dual parallel text encoding}.
\newblock In \emph{Proceedings of the 29th International Conference on Computational Linguistics, {COLING} 2022, Gyeongju, Republic of Korea, October 12-17, 2022}, pages 1511--1520. International Committee on Computational Linguistics.

\bibitem[{Chen et~al.(2022)Chen, Li, Qin, Lu, Lin, Chen, and Liang}]{DBLP:conf/emnlp/ChenLQLLCL22}
Jiaqi Chen, Tong Li, Jinghui Qin, Pan Lu, Liang Lin, Chongyu Chen, and Xiaodan Liang. 2022.
\newblock \href {https://doi.org/10.18653/V1/2022.EMNLP-MAIN.218} {Unigeo: Unifying geometry logical reasoning via reformulating mathematical expression}.
\newblock In \emph{Proceedings of the 2022 Conference on Empirical Methods in Natural Language Processing, {EMNLP} 2022, Abu Dhabi, United Arab Emirates, December 7-11, 2022}, pages 3313--3323. Association for Computational Linguistics.

\bibitem[{Chen et~al.(2021)Chen, Tang, Qin, Liang, Liu, Xing, and Lin}]{dblp:conf/acl/chentqllxl21}
Jiaqi Chen, Jianheng Tang, Jinghui Qin, Xiaodan Liang, Lingbo Liu, Eric~P. Xing, and Liang Lin. 2021.
\newblock \href {https://doi.org/10.18653/V1/2021.FINDINGS-ACL.46} {Geoqa: {A} geometric question answering benchmark towards multimodal numerical reasoning}.
\newblock In \emph{Findings of the Association for Computational Linguistics: {ACL/IJCNLP} 2021, Online Event, August 1-6, 2021}, volume {ACL/IJCNLP} 2021 of \emph{Findings of {ACL}}, pages 513--523. Association for Computational Linguistics.

\bibitem[{Chen et~al.(2023{\natexlab{a}})Chen, Li, Dong, Zhang, He, Wang, Zhao, and Lin}]{chen2023sharegpt4v}
Lin Chen, Jinsong Li, Xiaoyi Dong, Pan Zhang, Conghui He, Jiaqi Wang, Feng Zhao, and Dahua Lin. 2023{\natexlab{a}}.
\newblock \href {https://doi.org/10.48550/ARXIV.2311.12793} {Sharegpt4v: Improving large multi-modal models with better captions}.
\newblock \emph{CoRR}, abs/2311.12793.

\bibitem[{Chen et~al.(2023{\natexlab{b}})Chen, Wu, Wang, Su, Chen, Xing, Zhong, Zhang, Zhu, Lu, Li, Luo, Lu, Qiao, and Dai}]{chen2023internvl}
Zhe Chen, Jiannan Wu, Wenhai Wang, Weijie Su, Guo Chen, Sen Xing, Muyan Zhong, Qinglong Zhang, Xizhou Zhu, Lewei Lu, Bin Li, Ping Luo, Tong Lu, Yu~Qiao, and Jifeng Dai. 2023{\natexlab{b}}.
\newblock \href {https://doi.org/10.48550/ARXIV.2312.14238} {Internvl: Scaling up vision foundation models and aligning for generic visual-linguistic tasks}.
\newblock \emph{CoRR}, abs/2312.14238.

\bibitem[{Dai et~al.(2023)Dai, Li, Li, Tiong, Zhao, Wang, Li, Fung, and Hoi}]{instructblip}
Wenliang Dai, Junnan Li, Dongxu Li, Anthony Meng~Huat Tiong, Junqi Zhao, Weisheng Wang, Boyang Li, Pascale Fung, and Steven C.~H. Hoi. 2023.
\newblock \href {http://papers.nips.cc/paper\_files/paper/2023/hash/9a6a435e75419a836fe47ab6793623e6-Abstract-Conference.html} {Instructblip: Towards general-purpose vision-language models with instruction tuning}.
\newblock In \emph{Advances in Neural Information Processing Systems 36: Annual Conference on Neural Information Processing Systems 2023, NeurIPS 2023, New Orleans, LA, USA, December 10 - 16, 2023}.

\bibitem[{Dong et~al.(2024)Dong, Zhang, Zang, Cao, Wang, Ouyang, Wei, Zhang, Duan, Cao, Zhang, Li, Yan, Gao, Zhang, Li, Li, Chen, He, Zhang, Qiao, Lin, and Wang}]{internlmxcomposer2}
Xiaoyi Dong, Pan Zhang, Yuhang Zang, Yuhang Cao, Bin Wang, Linke Ouyang, Xilin Wei, Songyang Zhang, Haodong Duan, Maosong Cao, Wenwei Zhang, Yining Li, Hang Yan, Yang Gao, Xinyue Zhang, Wei Li, Jingwen Li, Kai Chen, Conghui He, Xingcheng Zhang, Yu~Qiao, Dahua Lin, and Jiaqi Wang. 2024.
\newblock \href {https://doi.org/10.48550/ARXIV.2401.16420} {Internlm-xcomposer2: Mastering free-form text-image composition and comprehension in vision-language large model}.
\newblock \emph{CoRR}, abs/2401.16420.

\bibitem[{Gao et~al.(2023)Gao, Pi, Zhang, Ye, Zhong, Wang, Hong, Han, Xu, Li, and Kong}]{gao2023gllava}
Jiahui Gao, Renjie Pi, Jipeng Zhang, Jiacheng Ye, Wanjun Zhong, Yufei Wang, Lanqing Hong, Jianhua Han, Hang Xu, Zhenguo Li, and Lingpeng Kong. 2023.
\newblock \href {https://doi.org/10.48550/ARXIV.2312.11370} {G-llava: Solving geometric problem with multi-modal large language model}.
\newblock \emph{CoRR}, abs/2312.11370.

\bibitem[{Gou et~al.(2023)Gou, Shao, Gong, Shen, Yang, Huang, Duan, and Chen}]{tora}
Zhibin Gou, Zhihong Shao, Yeyun Gong, Yelong Shen, Yujiu Yang, Minlie Huang, Nan Duan, and Weizhu Chen. 2023.
\newblock \href {https://doi.org/10.48550/ARXIV.2309.17452} {Tora: {A} tool-integrated reasoning agent for mathematical problem solving}.
\newblock \emph{CoRR}, abs/2309.17452.

\bibitem[{Hessel et~al.(2021)Hessel, Holtzman, Forbes, Bras, and Choi}]{DBLP:conf/emnlp/HesselHFBC21}
Jack Hessel, Ari Holtzman, Maxwell Forbes, Ronan~Le Bras, and Yejin Choi. 2021.
\newblock \href {https://doi.org/10.18653/V1/2021.EMNLP-MAIN.595} {Clipscore: {A} reference-free evaluation metric for image captioning}.
\newblock In \emph{Proceedings of the 2021 Conference on Empirical Methods in Natural Language Processing, {EMNLP} 2021, Virtual Event / Punta Cana, Dominican Republic, 7-11 November, 2021}, pages 7514--7528. Association for Computational Linguistics.

\bibitem[{Li et~al.(2023)Li, Yang, Liu, Ma, Zhang, Yang, Sun, Liu, and Bai}]{monkey}
Zhang Li, Biao Yang, Qiang Liu, Zhiyin Ma, Shuo Zhang, Jingxu Yang, Yabo Sun, Yuliang Liu, and Xiang Bai. 2023.
\newblock \href {https://doi.org/10.48550/ARXIV.2311.06607} {Monkey: Image resolution and text label are important things for large multi-modal models}.
\newblock \emph{CoRR}, abs/2311.06607.

\bibitem[{Lightman et~al.(2023)Lightman, Kosaraju, Burda, Edwards, Baker, Lee, Leike, Schulman, Sutskever, and Cobbe}]{stepbystep}
Hunter Lightman, Vineet Kosaraju, Yura Burda, Harrison Edwards, Bowen Baker, Teddy Lee, Jan Leike, John Schulman, Ilya Sutskever, and Karl Cobbe. 2023.
\newblock \href {https://doi.org/10.48550/ARXIV.2305.20050} {Let's verify step by step}.
\newblock \emph{CoRR}, abs/2305.20050.

\bibitem[{Liu et~al.(2023{\natexlab{a}})Liu, Guan, Li, Chen, Yacoob, Manocha, and Zhou}]{DBLP:journals/corr/abs-2310-14566}
Fuxiao Liu, Tianrui Guan, Zongxia Li, Lichang Chen, Yaser Yacoob, Dinesh Manocha, and Tianyi Zhou. 2023{\natexlab{a}}.
\newblock \href {https://doi.org/10.48550/ARXIV.2310.14566} {Hallusionbench: You see what you think? or you think what you see? an image-context reasoning benchmark challenging for gpt-4v(ision), llava-1.5, and other multi-modality models}.
\newblock \emph{CoRR}, abs/2310.14566.

\bibitem[{Liu et~al.(2023{\natexlab{b}})Liu, Li, Li, and Lee}]{liu2023improvedllava}
Haotian Liu, Chunyuan Li, Yuheng Li, and Yong~Jae Lee. 2023{\natexlab{b}}.
\newblock \href {https://doi.org/10.48550/ARXIV.2310.03744} {Improved baselines with visual instruction tuning}.
\newblock \emph{CoRR}, abs/2310.03744.

\bibitem[{Liu et~al.(2024)Liu, Li, Li, Li, Zhang, Shen, and Lee}]{liu2024llavanext}
Haotian Liu, Chunyuan Li, Yuheng Li, Bo~Li, Yuanhan Zhang, Sheng Shen, and Yong~Jae Lee. 2024.
\newblock \href {https://llava-vl.github.io/blog/2024-01-30-llava-next/} {Llava-next: Improved reasoning, ocr, and world knowledge}.

\bibitem[{Liu et~al.(2023{\natexlab{c}})Liu, Li, Wu, and Lee}]{liu2023llava}
Haotian Liu, Chunyuan Li, Qingyang Wu, and Yong~Jae Lee. 2023{\natexlab{c}}.
\newblock \href {http://papers.nips.cc/paper\_files/paper/2023/hash/6dcf277ea32ce3288914faf369fe6de0-Abstract-Conference.html} {Visual instruction tuning}.
\newblock In \emph{Advances in Neural Information Processing Systems 36: Annual Conference on Neural Information Processing Systems 2023, NeurIPS 2023, New Orleans, LA, USA, December 10 - 16, 2023}.

\bibitem[{Lu et~al.(2024{\natexlab{a}})Lu, Liu, Zhang, Wang, Dong, Liu, Sun, Ren, Li, Yang, Sun, Deng, Xu, Xie, and Ruan}]{deepseekvl}
Haoyu Lu, Wen Liu, Bo~Zhang, Bingxuan Wang, Kai Dong, Bo~Liu, Jingxiang Sun, Tongzheng Ren, Zhuoshu Li, Hao Yang, Yaofeng Sun, Chengqi Deng, Hanwei Xu, Zhenda Xie, and Chong Ruan. 2024{\natexlab{a}}.
\newblock \href {https://doi.org/10.48550/ARXIV.2403.05525} {Deepseek-vl: Towards real-world vision-language understanding}.
\newblock \emph{CoRR}, abs/2403.05525.

\bibitem[{Lu et~al.(2024{\natexlab{b}})Lu, Bansal, Xia, Liu, Li, Hajishirzi, Cheng, Chang, Galley, and Gao}]{lu2024mathvista}
Pan Lu, Hritik Bansal, Tony Xia, Jiacheng Liu, Chunyuan Li, Hannaneh Hajishirzi, Hao Cheng, Kai-Wei Chang, Michel Galley, and Jianfeng Gao. 2024{\natexlab{b}}.
\newblock Mathvista: Evaluating mathematical reasoning of foundation models in visual contexts.
\newblock In \emph{International Conference on Learning Representations (ICLR)}.

\bibitem[{Lu et~al.(2021)Lu, Gong, Jiang, Qiu, Huang, Liang, and Zhu}]{dblp:conf/acl/lugjqhlz20}
Pan Lu, Ran Gong, Shibiao Jiang, Liang Qiu, Siyuan Huang, Xiaodan Liang, and Song{-}Chun Zhu. 2021.
\newblock \href {https://doi.org/10.18653/V1/2021.ACL-LONG.528} {Inter-gps: Interpretable geometry problem solving with formal language and symbolic reasoning}.
\newblock In \emph{Proceedings of the 59th Annual Meeting of the Association for Computational Linguistics and the 11th International Joint Conference on Natural Language Processing, {ACL/IJCNLP} 2021, (Volume 1: Long Papers), Virtual Event, August 1-6, 2021}, pages 6774--6786. Association for Computational Linguistics.

\bibitem[{Luo et~al.(2023)Luo, Sun, Xu, Zhao, Lou, Tao, Geng, Lin, Chen, and Zhang}]{luo2023wizardmath}
Haipeng Luo, Qingfeng Sun, Can Xu, Pu~Zhao, Jianguang Lou, Chongyang Tao, Xiubo Geng, Qingwei Lin, Shifeng Chen, and Dongmei Zhang. 2023.
\newblock \href {https://doi.org/10.48550/ARXIV.2308.09583} {Wizardmath: Empowering mathematical reasoning for large language models via reinforced evol-instruct}.
\newblock \emph{CoRR}, abs/2308.09583.

\bibitem[{Masry et~al.(2022)Masry, Long, Tan, Joty, and Hoque}]{dblp:conf/acl/masryltjh22}
Ahmed Masry, Do~Xuan Long, Jia~Qing Tan, Shafiq~R. Joty, and Enamul Hoque. 2022.
\newblock \href {https://doi.org/10.18653/V1/2022.FINDINGS-ACL.177} {Chartqa: {A} benchmark for question answering about charts with visual and logical reasoning}.
\newblock In \emph{Findings of the Association for Computational Linguistics: {ACL} 2022, Dublin, Ireland, May 22-27, 2022}, pages 2263--2279. Association for Computational Linguistics.

\bibitem[{OpenAI(2023{\natexlab{a}})}]{openai2023gpt4}
OpenAI. 2023{\natexlab{a}}.
\newblock \href {https://doi.org/10.48550/ARXIV.2303.08774} {{GPT-4} technical report}.
\newblock \emph{CoRR}, abs/2303.08774.

\bibitem[{OpenAI(2023{\natexlab{b}})}]{openai2023gpt4v}
OpenAI. 2023{\natexlab{b}}.
\newblock \href {https://api.semanticscholar.org/CorpusID:263218031} {Gpt-4v(ision) system card}.
\newblock In \emph{technical report}.

\bibitem[{Shao et~al.(2024)Shao, Wang, Zhu, Xu, Song, Zhang, Li, Wu, and Guo}]{DBLP:journals/corr/abs-2402-03300}
Zhihong Shao, Peiyi Wang, Qihao Zhu, Runxin Xu, Junxiao Song, Mingchuan Zhang, Y.~K. Li, Y.~Wu, and Daya Guo. 2024.
\newblock \href {https://doi.org/10.48550/ARXIV.2402.03300} {Deepseekmath: Pushing the limits of mathematical reasoning in open language models}.
\newblock \emph{CoRR}, abs/2402.03300.

\bibitem[{Toshniwal et~al.(2024)Toshniwal, Moshkov, Narenthiran, Gitman, Jia, and Gitman}]{openmath}
Shubham Toshniwal, Ivan Moshkov, Sean Narenthiran, Daria Gitman, Fei Jia, and Igor Gitman. 2024.
\newblock \href {https://doi.org/10.48550/ARXIV.2402.10176} {Openmathinstruct-1: {A} 1.8 million math instruction tuning dataset}.
\newblock \emph{CoRR}, abs/2402.10176.

\bibitem[{Wang et~al.(2024)Wang, Pan, Shi, Lu, Zhan, and Li}]{wang2024measuring}
Ke~Wang, Junting Pan, Weikang Shi, Zimu Lu, Mingjie Zhan, and Hongsheng Li. 2024.
\newblock \href {https://doi.org/10.48550/ARXIV.2402.14804} {Measuring multimodal mathematical reasoning with math-vision dataset}.
\newblock \emph{CoRR}, abs/2402.14804.

\bibitem[{Wang et~al.(2023{\natexlab{a}})Wang, Ren, Zhou, Lu, Luo, Shi, Zhang, Song, Zhan, and Li}]{mathcoder}
Ke~Wang, Houxing Ren, Aojun Zhou, Zimu Lu, Sichun Luo, Weikang Shi, Renrui Zhang, Linqi Song, Mingjie Zhan, and Hongsheng Li. 2023{\natexlab{a}}.
\newblock \href {https://doi.org/10.48550/ARXIV.2310.03731} {Mathcoder: Seamless code integration in llms for enhanced mathematical reasoning}.
\newblock \emph{CoRR}, abs/2310.03731.

\bibitem[{Wang et~al.(2023{\natexlab{b}})Wang, Li, Shao, Xu, Dai, Li, Chen, Wu, and Sui}]{DBLP:journals/corr/abs-2312-08935}
Peiyi Wang, Lei Li, Zhihong Shao, R.~X. Xu, Damai Dai, Yifei Li, Deli Chen, Y.~Wu, and Zhifang Sui. 2023{\natexlab{b}}.
\newblock \href {https://doi.org/10.48550/ARXIV.2312.08935} {Math-shepherd: Verify and reinforce llms step-by-step without human annotations}.
\newblock \emph{CoRR}, abs/2312.08935.

\bibitem[{Wei et~al.(2022)Wei, Wang, Schuurmans, Bosma, Ichter, Xia, Chi, Le, and Zhou}]{DBLP:conf/nips/Wei0SBIXCLZ22}
Jason Wei, Xuezhi Wang, Dale Schuurmans, Maarten Bosma, Brian Ichter, Fei Xia, Ed~H. Chi, Quoc~V. Le, and Denny Zhou. 2022.
\newblock \href {http://papers.nips.cc/paper\_files/paper/2022/hash/9d5609613524ecf4f15af0f7b31abca4-Abstract-Conference.html} {Chain-of-thought prompting elicits reasoning in large language models}.
\newblock In \emph{Advances in Neural Information Processing Systems 35: Annual Conference on Neural Information Processing Systems 2022, NeurIPS 2022, New Orleans, LA, USA, November 28 - December 9, 2022}.

\bibitem[{Xu et~al.(2020)Xu, Zhang, Mao, Wang, Xie, and Zhang}]{DBLP:conf/acl/XuZMWXZ20}
Benfeng Xu, Licheng Zhang, Zhendong Mao, Quan Wang, Hongtao Xie, and Yongdong Zhang. 2020.
\newblock \href {https://doi.org/10.18653/V1/2020.ACL-MAIN.542} {Curriculum learning for natural language understanding}.
\newblock In \emph{Proceedings of the 58th Annual Meeting of the Association for Computational Linguistics, {ACL} 2020, Online, July 5-10, 2020}, pages 6095--6104. Association for Computational Linguistics.

\bibitem[{Yao et~al.(2022)Yao, Huang, Hou, Lu, Niu, Xu, Liang, Li, Jiang, and Xu}]{DBLP:conf/iclr/YaoHHLNXLLJX22}
Lewei Yao, Runhui Huang, Lu~Hou, Guansong Lu, Minzhe Niu, Hang Xu, Xiaodan Liang, Zhenguo Li, Xin Jiang, and Chunjing Xu. 2022.
\newblock \href {https://openreview.net/forum?id=cpDhcsEDC2} {{FILIP:} fine-grained interactive language-image pre-training}.
\newblock In \emph{The Tenth International Conference on Learning Representations, {ICLR} 2022, Virtual Event, April 25-29, 2022}. OpenReview.net.

\bibitem[{Ye et~al.(2023)Ye, Xu, Ye, Yan, Hu, Liu, Qian, Zhang, Huang, and Zhou}]{mplug2}
Qinghao Ye, Haiyang Xu, Jiabo Ye, Ming Yan, Anwen Hu, Haowei Liu, Qi~Qian, Ji~Zhang, Fei Huang, and Jingren Zhou. 2023.
\newblock \href {https://doi.org/10.48550/ARXIV.2311.04257} {mplug-owl2: Revolutionizing multi-modal large language model with modality collaboration}.
\newblock \emph{CoRR}, abs/2311.04257.

\bibitem[{Yu et~al.(2023)Yu, Jiang, Shi, Yu, Liu, Zhang, Kwok, Li, Weller, and Liu}]{yu2023metamath}
Longhui Yu, Weisen Jiang, Han Shi, Jincheng Yu, Zhengying Liu, Yu~Zhang, James~T. Kwok, Zhenguo Li, Adrian Weller, and Weiyang Liu. 2023.
\newblock \href {https://doi.org/10.48550/ARXIV.2309.12284} {Metamath: Bootstrap your own mathematical questions for large language models}.
\newblock \emph{CoRR}, abs/2309.12284.

\bibitem[{Yue et~al.(2023)Yue, Ni, Zhang, Zheng, Liu, Zhang, Stevens, Jiang, Ren, Sun, Wei, Yu, Yuan, Sun, Yin, Zheng, Yang, Liu, Huang, Sun, Su, and Chen}]{mmmu}
Xiang Yue, Yuansheng Ni, Kai Zhang, Tianyu Zheng, Ruoqi Liu, Ge~Zhang, Samuel Stevens, Dongfu Jiang, Weiming Ren, Yuxuan Sun, Cong Wei, Botao Yu, Ruibin Yuan, Renliang Sun, Ming Yin, Boyuan Zheng, Zhenzhu Yang, Yibo Liu, Wenhao Huang, Huan Sun, Yu~Su, and Wenhu Chen. 2023.
\newblock \href {https://doi.org/10.48550/ARXIV.2311.16502} {{MMMU:} {A} massive multi-discipline multimodal understanding and reasoning benchmark for expert {AGI}}.
\newblock \emph{CoRR}, abs/2311.16502.

\bibitem[{Zhao et~al.(2023)Zhao, Xie, Kawaguchi, He, and Xie}]{DBLP:conf/emnlp/ZhaoXKHX23}
James~Xu Zhao, Yuxi Xie, Kenji Kawaguchi, Junxian He, and Michael~Qizhe Xie. 2023.
\newblock \href {https://doi.org/10.18653/V1/2023.FINDINGS-EMNLP.55} {Automatic model selection with large language models for reasoning}.
\newblock In \emph{Findings of the Association for Computational Linguistics: {EMNLP} 2023, Singapore, December 6-10, 2023}, pages 758--783. Association for Computational Linguistics.

\bibitem[{Zhou et~al.(2023{\natexlab{a}})Zhou, Wang, Lu, Shi, Luo, Qin, Lu, Jia, Song, Zhan, and Li}]{DBLP:journals/corr/abs-2308-07921}
Aojun Zhou, Ke~Wang, Zimu Lu, Weikang Shi, Sichun Luo, Zipeng Qin, Shaoqing Lu, Anya Jia, Linqi Song, Mingjie Zhan, and Hongsheng Li. 2023{\natexlab{a}}.
\newblock \href {https://doi.org/10.48550/ARXIV.2308.07921} {Solving challenging math word problems using {GPT-4} code interpreter with code-based self-verification}.
\newblock \emph{CoRR}, abs/2308.07921.

\bibitem[{Zhou et~al.(2023{\natexlab{b}})Zhou, Liu, Xu, Iyer, Sun, Mao, Ma, Efrat, Yu, Yu, Zhang, Ghosh, Lewis, Zettlemoyer, and Levy}]{DBLP:conf/nips/ZhouLX0SMMEYYZG23}
Chunting Zhou, Pengfei Liu, Puxin Xu, Srinivasan Iyer, Jiao Sun, Yuning Mao, Xuezhe Ma, Avia Efrat, Ping Yu, Lili Yu, Susan Zhang, Gargi Ghosh, Mike Lewis, Luke Zettlemoyer, and Omer Levy. 2023{\natexlab{b}}.
\newblock \href {http://papers.nips.cc/paper\_files/paper/2023/hash/ac662d74829e4407ce1d126477f4a03a-Abstract-Conference.html} {{LIMA:} less is more for alignment}.
\newblock In \emph{Advances in Neural Information Processing Systems 36: Annual Conference on Neural Information Processing Systems 2023, NeurIPS 2023, New Orleans, LA, USA, December 10 - 16, 2023}.

\end{thebibliography}

\appendix

\appendix

\section{Detailed Task Information}
\label{sec:Detailed_TI}
Table~\ref{tab:geometry_fields_sources} shows the correspondence between abbreviations and detailed task names.

\section{Training Parameters}
\label{sec:training_parameters}
We keep the same parameters as those specified by the model's original authors.
Detail parameters are shown in Table~\ref{tab:appendix_param}.

\section{Prompts}
\subsection{Prompt for Question-Answer Pairs Generation}
\label{sec:qa_prompt}
Table~\ref{tab:qa_prompt} shows the prompt for question-answer pairs generation.
We prompt GPT-4V to generate simplified geometric problems based on the open-source datasets.
\subsection{Prompt for Wolfram Code Generation}
\label{sec:code_prompt}
Table~\ref{tab:code_prompt} shows the prompt for Wolfram code generation.
We prompt GPT-4 to generate the Wolfram code based on the information from the question, the answer, and the image description.

\subsection{Prompt for Scoring}
\label{sec:scoring_prompt}
Table~\ref{tab:scoring_prompt} shows the prompt for scoring.
We prompt GPT-4V to score the degree of alignment between the images and the questions.

\subsection{Prompt for Difficulty Comparison}
\label{sec:difficulty_prompt}
Table~\ref{tab:difficulty_prompt} shows the prompt for difficulty comparison.
We employ GPT-4V to determine which of the two problems is more difficult.

\section{Detailed Evaluation Results}
\label{sec:detail_results}

\subsection{MathVista Results}
\label{sec:detail_mathvista}
We show full MathVista testmini results in Table~\ref{tab:appendix_mathvista}.
Although our method focuses on geometric problems, the GeoGPT4V dataset can still improve the overall scores of various models, except InternVL-Chat-V1.2-Plus, which has already 
employed a customized fine-tuning dataset with 12 million samples.

\subsection{MathVision Results}
\label{sec:detail_mathvision}
We show full MathVision test results in Table~\ref{tab:appendix_mathvision}.
We can find that the GeoGPT4V dataset can improve the scores of most tasks on MathVision for various models.
The results demonstrate the effectiveness of the GeoGPT4V dataset.

\subsection{Comparison with Other Models.}
\label{sec:comparsion_other_models}
We compare the performance of our best model, InternVL-Chat-V1.2-Plus-GeoGPT4V, with other open-source and closed-source models regarding geometric capabilities.
Detailed results are in Table~\ref{tab:appendix_comparison}.

For MathVista, our best model achieves the best geometric scores among all models.
For MathVision, our best model achieves the highest scores for average score and most geometric scores among open-source models.
The experimental results demonstrate the effectiveness of the GeoGPT4V dataset.

\begin{table}[htbp]
    \centering
     \renewcommand\arraystretch{0.9} 
    \resizebox{0.5\textwidth}{!}{
    \begin{tabular}{ll}
\toprule
        \textbf{Abbreviation} & \textbf{Task} \\
\toprule

\multicolumn{2}{l}{\textit{\textbf{MathVista}}} \\ 
\midrule
FQA & Figure Question Answering \\
GPS & Geometry Problem Solving \\
MWP & Math Word Problem \\
TQA & Textbook question answering \\
VQA & Visual Question Answering \\
ALG & Algebraic Reasoning \\
ARI & Arithmetic Reasoning \\
GEO & Geometry Reasoning \\
LOG & Logical Reasoning \\
NUM & Numeric Commonsense \\
SCI & Scientific Reasoning \\
STA & Statistical Reasoning \\
\midrule

\multicolumn{2}{l}{\textit{\textbf{MathVision}}} \\ 
\midrule
Alg & Algebra \\
AnaG & Analytic Geometry \\
Ari & Arithmetic \\
CombG & Combinatorial Geometry \\
Comb & Combinatorics\\
Cnt & Counting \\
DescG & Descriptive Geometry \\
GrphT & Graph Theory \\
Log & Logic \\
Angle & Metric Geometry - Angle \\
Area & Metric Geometry - Area \\
Len & Metric Geometry - Length \\
SolG & Solid Geometry \\
Stat & Statistics \\
Topo & Topology \\
TransG & Transformation Geometry \\
\bottomrule
    \end{tabular}}
    \caption{\textbf{Correspondence between abbreviations and detailed task names in MathVista and MathVision benchmarks.}}
    \label{tab:geometry_fields_sources}
\end{table}

\begin{table*}[htbp]
\centering
\resizebox{0.8\textwidth}{!}{
\begin{tabular}{l|ccc}
\toprule
Parameters & LLaVA-1.5 & ShareGPT4V & InternVL-Chat-V1.2-Plus \\ 
\toprule
Train Epochs & 1 & 1 & 1\\
Global Batch Size & 128 & 128 & 128 \\
Learning Rate & $2e^{-5}$ & $2e^{-5}$ & $1e^{-5}$\\
Learning Rate Schedule & cosine decay & cosine decay & cosine decay\\
Weight Decay & 0 & 0 & 0.05\\
Warmup Ratio & 0.03 & 0.03 & 0.03\\
Optimizer & AdamW & AdamW & AdamW \\
Tune Visual Encoder & False & False & False \\
Tune MLP & True & True & True \\
Tune LLM & True & True & True \\ 
\bottomrule

\end{tabular}
}
\caption{\textbf{Training parameters of different models.}
To make a fair comparison, we keep the training parameters consistent with those specified by the model's original authors and train the models for one epoch.
}
\label{tab:appendix_param}
\end{table*}
\begin{table*}[htbp]
\centering
\begin{minipage}{1\textwidth} 
\centering
\begin{tcolorbox} 
\small

\begin{tabular}{p{\columnwidth}}
Please act as a question generator.

Give you a question and its answer, along with a corresponding image for the question; please generate new questions and provide new answers in English. 
The new questions and new answers must meet the following conditions:

1. The new questions are slightly easier than the original ones but shouldn't be too simple.

2. Do not merely rephrase the question; you must reduce its difficulty level.

3. The new question must include a detailed description of the information in the image, which must be detailed enough to allow others to redraw the image based on the description.

5. The questions should be as diverse as possible.

6. The new answers must be correct.

Some useful tips: 

1. You can incorporate information from the original answer into the question.

3. You can generate lead-up problems for the original problem.

5. You can generate sub-problems for the original problem.

4. Imagine that others cannot see the image corresponding to the new question; you must describe it using words.

5. For each question, consider it as a standalone item. Others can only view one question at a time, so avoid using phrases like "similar to the previous question" or references such as "New\_Image 1".

Come up with three diverse questions and answers.

Input format:

Question: <question example>

Answer: <answer example>

You must follow this output format: 

New\_Question: <new question example> 

New\_Answer: <new answer example>

Image\_Description: <new image description example>

\end{tabular}
\end{tcolorbox}

\caption{\textbf{Prompt for Question-Answer Pairs Generation.} 
We prompt GPT-4V to generate simplified questions.
We also prompt GPT-4V to generate questions that are as diverse as possible to prevent GPT-4V from generating the same questions.
}
\label{tab:qa_prompt}
\end{minipage}
\end{table*}
\begin{table*}[htbp]
\centering
\begin{minipage}{1\textwidth} 
\centering
\begin{tcolorbox} 
\small

\begin{tabular}{p{\columnwidth}}
You are a teacher creating an exam, and you need to draw images for the questions on the exam.

Give you a question, an answer, and an image description, and generate the image corresponding to the question using Mathematica code. 
Your code must meet the following conditions:

1. Only use the ``Export'' command at the end of the code to save the generated image to ``/temp/image.png''.

2. The image should be clear and correspond to the question, with particular attention to shape and angle.

3. You only need to generate the image; there is no need to solve the problem.

4. All variables in the code should be named for easy understanding; avoid using terms such as ``C'' directly.

Some useful tips:

1. Focus on the image description.

2. You can use the information from the question and answer to help you generate code.

Come up with one code.

Input format:

Question: <question example>

Answer: <answer example>

Image description: <image description example>

You must follow this output format:

Code: <code example>

\end{tabular}
\end{tcolorbox}
\caption{\textbf{Prompt for Wolfram Code Generation.} 
When prompting GPT-4, we integrate both image descriptions and question-answer data to refine code generation. 
Additionally, we prompt GPT-4 to ensure variable naming within the code for clarity, aiming to enhance GPT-4's grasp of the code's relationship to the query at hand.
}
\label{tab:code_prompt}
\end{minipage}
\end{table*}
\begin{table*}[htbp]
\centering
\begin{minipage}{1\textwidth} 
\centering
\begin{tcolorbox} 
\small

\begin{tabular}{p{\columnwidth}}
Please act as a scorer.

Give you a description, along with an image. Please evaluate the degree of match between the image and the description and give a score. 
The evaluation process must meet the following conditions:

1. The score is a decimal between 0 and 1.

2. The score reflects the degree of image-description match.

3. If the image and the image description do not match, the score should be low.

4. The score should be lower if the image is not clear enough or difficult to understand.

5. The image should be rated low if it contains only text and numbers, with no geometric shapes or chart forms.

6. The image must have clear shapes and labels.

Some useful tips: 

1. Don't always give high scores.

2. Only give high scores when the image and the description match very well.

3. You can use two decimal places to represent your score.

Come up with one score.

Input format:

Image description: <image description example>

You must follow this output format: 

Reason: <your reason example>

Score: <score example>

\end{tabular}
\end{tcolorbox}
\caption{\textbf{Prompt for Scoring.} 
We employ GPT-4V to score the degree of alignment between the generated images and the questions.
Specifically, the score is a decimal that ranges from 0 to 1. 
We also prompt GPT-4V to give a reason first and then give a final score, hoping this can enhance the accuracy of scoring.
}
\label{tab:scoring_prompt}
\end{minipage}
\end{table*}
\begin{table*}[htbp]
\centering
\begin{minipage}{1\textwidth} 
\centering
\begin{tcolorbox} 
\small

\begin{tabular}{p{\columnwidth}}
Please act as a difficulty level evaluator.

Give two geometric data, each consisting of a question, an answer, and an image.

Please compare these two questions to determine which one is more difficult.

If the first one is more difficult, output ``1''; if the second one is more difficult, output ``2''.

Some useful tips: 

1. You should consider the complexity and difficulty of the questions and images.

2. Don't automatically assume that multiple-choice questions are easier.

3. A shorter answer does not mean it's easier.

Input format:

Question\_1: <the first question>

Answer\_1: <the first answer>

Question\_2: <the second question>

Answer\_2: <the second answer>

The first image corresponds to the first question, and the second image corresponds to the second question.

You can only output the number ``1'' or ``2''.

\end{tabular}
\end{tcolorbox}

\caption{\textbf{Prompt for Difficulty Comparison.} 
We prompt GPT-4V to determine which of the two questions is more difficult.
We instruct GPT-4V not to simplistically assume that multiple-choice questions or shorter answers imply an easier question.
}
\label{tab:difficulty_prompt}
\end{minipage}
\end{table*}
\begin{table*}[]
\small
 \renewcommand\tabcolsep{1.8pt} 
 \renewcommand\arraystretch{1.2} 
\centering

\begin{tabular}{lr|c|ccccc|ccccccc}
\toprule
Model & Size & All & FQA & GPS & MWP & TQA & VQA & ALG & ARI & GEO & LOG & NUM & SCI & STA\\ 
\toprule
LLaVA-1.5 & 7B & 25.1$^{*}$&23.79$^{*}$&20.67$^{*}$&12.90$^{*}$&39.24$^{*}$&32.40$^{*}$&24.20$^{*}$&22.10$^{*}$&20.92$^{*}$&16.22$^{*}$&18.75$^{*}$&	36.89$^{*}$&22.26$^{*}$	\\
LLaVA-1.5 & 13B & 27.3$^{*}$ & 22.68$^{*}$ & 24.04$^{*}$ & 16.67$^{*}$ & 42.41$^{*}$ & 35.75$^{*}$ & 27.40$^{*}$ & 24.93$^{*}$ & 23.85$^{*}$ & 18.92$^{*}$ & 25.00$^{*}$ & 39.34$^{*}$ & 22.59$^{*}$\\
\midrule
LLaVA-1.5-G & 7B & \color{up}30.7&\color{up}28.25 & \color{up}32.69 & \color{up}18.28 & \color{up}42.41 & \color{up}34.64 & \color{up}32.38 & \color{up}25.78 & \color{up}32.22 & \color{up}32.43 & \color{up}23.61 & \color{up}42.62 & \color{up}26.58\\
LLaVA-1.5-G & 13B & \color{up}32.2 & \color{up}28.25 & \color{up}36.54 & \color{up}19.89 & \color{down}41.14 & \color{up}37.99 & \color{up}35.23 & \color{up}28.05 & \color{up}37.24 & \color{up}27.03 & \color{up}26.39 & \color{up}42.62 & \color{up}27.57\\ 
\midrule
ShareGPT4V & 7B & 27.3$^{*}$ & 21.93$^{*}$ & 21.63$^{*}$ & 19.35$^{*}$ & 43.04$^{*}$ & 36.31$^{*}$ & 24.91$^{*}$ & 27.20$^{*}$ & 20.50$^{*}$ & 18.92$^{*}$ & 22.92$^{*}$ & 40.16$^{*}$ & 21.93$^{*}$\\
ShareGPT4V & 13B & 30.4$^{*}$ & 23.97$^{*}$ & 27.40$^{*}$ & 25.81$^{*}$ & 43.67$^{*}$ & 36.87$^{*}$ & 28.83$^{*}$ & 31.16$^{*}$ & 27.62$^{*}$ & 10.81$^{*}$ & 26.39$^{*}$ & 41.80$^{*}$ & 26.91$^{*}$\\
\midrule
ShareGPT4V-G & 7B & \color{up}30.4 & \color{up}26.77 & \color{up}32.69 & \color{up}20.97 & \color{down}40.51 & \color{down}34.08 & \color{up}31.67 & \color{down}26.91 & \color{up}31.80 & \color{up}21.62 & \color{down}20.83 & \color{up}40.98 & \color{up}25.52\\
ShareGPT4V-G & 13B & \color{up}34.1 & \color{up}27.51 & \color{up}43.27 & \color{down}23.12 & \color{down}43.04 & 36.87 & \color{up}39.86 & \color{down}29.18 & \color{up}42.26 & \color{up}27.03 & \color{down}24.31 & \color{up}44.26 & \color{up}27.57\\ 
\midrule
InternVL\dag & 40B & 59.9 & 51.7 & 61.1 & 79.6 & 52.5 & 57.0 & 54.5 & 63.2 & 61.1 & 16.2 & 48.6 & 55.7 & 60.8 \\
InternVL-G\dag & 40B & \color{down}56.2 & \color{down}46.10 & \color{up}64.42 & \color{down}75.27 &\color{down}51.90 & \color{down}45.81 & \color{up}57.30 & \color{down}54.96 & \color{up}63.60 & \color{up}18.92 & \color{down}39.58 & \color{down}53.28 & \color{down}55.81\\ 
\bottomrule
\end{tabular}
\caption{\textbf{Overall results of different models on the MathVista.}
For the model trained with GeoGPT4V, score increases are marked in \textcolor{up}{red} 
compared to the original model.
$^{*}$ indicates our re-implemented test results missed in benchmarks or origin papers.
InternVL\dag represents the abbreviation for InternVL-Chat-V1.2-Plus.
The suffix “-G” to the model name indicates a model trained on the GeoGPT4V.
We present the detailed score for all the tasks such as “FQA” and “GPS”, as well as the overall (All) score for the benchmark.
Due to limited space, we utilize abbreviations for the tasks and illustrate the detailed task name in the Appendix \ref{sec:Detailed_TI}.
}
\label{tab:appendix_mathvista}
\end{table*}
\begin{table*}[]
\centering
\small
 \renewcommand\tabcolsep{1.5pt} 
 \renewcommand\arraystretch{1} 
\resizebox{\textwidth}{!}{
\begin{tabular}{lr|c|cccccccccccccccc}
\toprule
Model & Size & All & Alg & AnaG & Ari & CombG & Comb & Cnt & DescG & GrphT & Log & Angle & Area & Len & SolG & Stat & Topo & TransG \\ 
\toprule
LLaVA-1.5 & 7B & 8.52 & 7.0 & 7.1 & 10.7 & 7.1 & 4.8 & 10.5 & 7.7 & 10.0 & 9.2 & 15.6 & 10.2 & 9.8 & 5.3 & 8.6 & 4.4 & 4.8\\
LLaVA-1.5 & 13B & 11.12 & 7.0 & 14.3 & 14.3 & 9.1 & 6.6 & 6.0 & 13.5 & 5.6 & 13.5 & 10.4 & 12.6 & 14.7 & 11.5 & 13.8 & 13.0 & 10.7\\
\midrule
LLaVA-1.5-G & 7B & \color{up}12.89 & \color{up}8.41 & \color{up}9.52 & \color{down}9.29 & \color{up}16.88 & \color{up}6.55 & \color{down}10.45 & \color{up}9.62 & \color{up}21.11 & \color{up}12.61 & \color{up}19.08 & \color{up}11.06 & \color{up}17.15 & \color{up}9.43 & \color{up}13.79 & \color{up}13.04 & \color{up}15.48\\
LLaVA-1.5-G & 13B & \color{up}13.98 & \color{up}9.28 & \color{up}15.48 & \color{up}16.43 & \color{up}14.29 & \color{up}10.71 & \color{up}10.45 & \color{down}12.50 & \color{up}18.89 & \color{down}11.76 & \color{up}19.65 & \color{up}13.6 & \color{up}18.49 & \color{down}10.25 & \color{down}13.79 & \color{up}17.39 & \color{up}13.10\\ 
\midrule
ShareGPT4V & 7B & 10.53 & 5.5 & 3.6 & 12.9 & 10.1 & 4.8 & 7.5 & 11.5 & 14.4 & 10.9 & 16.2 & 11.8 & 12.3 & 9.8 & 15.5 & 17.4 & 11.3\\ 
ShareGPT4V & 13B & 11.88 & 7.5 & 15.5 & 16.4 & 10.7 & 8.9 & 9.0 & 11.5 & 8.9 & 7.6 & 11.6 & 13.0 & 17.4 & 10.3 & 8.6 & 8.7 & 12.5\\
\midrule
ShareGPT4V-G & 7B & \color{up}12.80 & \color{up}7.83 & \color{up}11.9 & \color{up}15.00 & \color{up}12.99 & \color{up}5.95 & \color{down}7.46 & \color{down}9.62 & \color{up}16.67 & \color{up}15.97 & \color{up}17.34 & \color{up}13.60 & \color{up}17.59 & \color{up}10.25 & \color{up}15.52 & \color{down}8.70 & \color{up}11.31\\
ShareGPT4V-G & 13B & \color{up}12.63 & \color{up}8.41 & \color{up}22.62 & \color{down}15.00 & \color{down}9.74 & \color{down}6.55 & \color{down}8.96 & \color{up}13.46 & \color{up}11.11 & \color{up}15.13 & \color{up}19.08 & \color{up}15.80 & \color{down}13.81 & \color{down}9.02 & \color{down}6.90 & \color{up}13.04 & \color{up}13.69\\ 
\midrule
InternVL\dag & 40B & 9.18$^{*}$ & 8.41$^{*}$ & 16.67$^{*}$ & 8.57$^{*}$ & 12.99$^{*}$ & 9.52$^{*}$ & 10.45$^{*}$ & 15.38$^{*}$ & 13.33$^{*}$ & 11.76$^{*}$ & 4.62$^{*}$ & 5.60$^{*}$ & 6.46$^{*}$ & 9.84$^{*}$ & 12.07$^{*}$ & 21.74$^{*}$ & 10.71$^{*}$\\
InternVL-G\dag & 40B & \color{up}16.12 & \color{up}9.57 & 16.67 & \color{up}15.00 & \color{up}18.18 & \color{up}10.71 & 10.45 & \color{down}13.46 & \color{up}16.67 & \color{up}16.81 & \color{up}23.12 & \color{up}18.4 & \color{up}18.93 & \color{up}11.89 & \color{down}6.90 & \color{down}13.04 & \color{up}23.21\\ 
\bottomrule
\end{tabular}
}
\caption{\textbf{Overall results of different models on the MathVision.}
For the model trained with GeoGPT4V, score increases are marked in \textcolor{up}{red} 
compared to the original model.
$^{*}$ indicates our re-implemented test results missed in benchmarks or origin papers.
InternVL\dag represents the abbreviation for InternVL-Chat-V1.2-Plus.
The suffix “-G” to the model name indicates a model trained on the GeoGPT4V.
We present the detailed score for all the tasks such as “Alg” and “AnaG”, as well as the overall (All) score for the benchmark.
Due to limited space, we utilize abbreviations for the tasks and illustrate the detailed task name in the Appendix \ref{sec:Detailed_TI}.
}
\label{tab:appendix_mathvision}
\end{table*}
\definecolor{my_blue}{HTML}{FFA8AB} 
\begin{table*}[]

\centering
\small
 \renewcommand\tabcolsep{2pt} 
 \renewcommand\arraystretch{1.2} 
\begin{tabular}{lr|ccc|cccccccccc}
\toprule
\multicolumn{1}{l}{\multirow{2}{*}{Model}} & \multicolumn{1}{r|}{\multirow{2}{*}{Size}} & \multicolumn{3}{c|}{MathVista}                                 & \multicolumn{10}{c}{MathVision}                                                                                                                                                                                                                                            \\ \cline{3-15} 
\multicolumn{1}{c}{}                       & \multicolumn{1}{c|}{}                       & GPS   & GEO &AVG   & AnaG  & CombG & DescG & GrphT & Angle & Area  & Len   & SolG  & TransG & AVG \\ 
\toprule

InternVL-G\dag                                   & 40B                                         & \cellcolor[HTML]{FFA8AB}\textbf{64.42}  & \cellcolor[HTML]{FFA8AB}\textbf{63.6}&\cellcolor[HTML]{FFA8AB}\textbf{64.01}  & \cellcolor[HTML]{FFA8AB}16.67  & \cellcolor[HTML]{FFA8AB}18.18  & 13.46  & 16.67   & \cellcolor[HTML]{FFA8AB}\textbf{23.12}  & \cellcolor[HTML]{FFA8AB}18.40    & \cellcolor[HTML]{FFA8AB}18.93  & 11.89  &\cellcolor[HTML]{FFA8AB} 23.21 &\cellcolor[HTML]{FFA8AB}17.84  \\ 
\midrule
\multicolumn{15}{c}{Open-source Models}\\ 
\midrule
LLaVA-1.5 & 13B & 24.04$^{*}$ & 23.85$^{*}$&23.95$^{*}$ & 14.3  & 9.1   & 13.5  & 5.6   & 10.4  & 12.6  & 14.7  & 11.5  & 10.7 &11.38\\
ShareGPT4V & 13B & 27.4$^{*}$  & 27.62$^{*}$&27.51$^{*}$ & 15.5  & 10.7  & 11.5  & 8.9   & 11.6  & 13    & 17.4  & 10.3  & 12.5&12.38 \\
G-LLaVA\ddag & 13B  & 56.25$^{*}$& 51.88$^{*}$&54.07$^{*}$ & 9.52$^{*}$& 7.79$^{*}$& 8.65$^{*}$&7.78$^{*}$&8.67$^{*}$&12.20$^{*}$&10.02$^{*}$&7.38$^{*}$&8.93$^{*}$&8.99$^{*}$\\
InternLM-VL\dag & 7B&63.0&62.3&62.65&15.5&15.3&14.4&\cellcolor[HTML]{FFA8AB}\textbf{22.2}&19.7&15.6&15.0&\cellcolor[HTML]{FFA8AB}11.9&15.5&16.12 \\ 
InternVL\dag& 40B & 61.1  & 61.1&61.1 & 16.67$^{*}$  & 12.99$^{*}$  & \cellcolor[HTML]{FFA8AB}15.38$^{*}$  & 13.33$^{*}$   & 4.62$^{*}$  & 5.60$^{*}$    & 6.46$^{*}$  & 9.84$^{*}$  & 10.71$^{*}$  &10.62$^{*}$ \\ 
\midrule
\multicolumn{15}{c}{Closed-source Models}\\ 
\midrule
Qwen-VL-Plus & -&38.5&39.3&38.90&17.9&12.7&15.4&8.9&11.6&6.4&10.0&14.3&11.31&12.06 \\
Qwen-VL-Max & -&-&-&-&19.1&16.9&16.4&12.2&13.3&14.2&19.8&11.5&17.3&15.61 \\
Gemini-1.0-Pro & -&40.4&41.0&40.70& 10.7&20.1&20.2&21.1&19.1&19.0&20.0&14.3&20.8&18.37\\
Gemini-1.0-Ultra & -&56.2&55.6&55.90&-&-&-&-&-&-&-&-&-&- \\
GPT-4V & -&50.5&51.0&50.75&\textbf{32.1}&\textbf{21.1}&\textbf{22.1}&14.4&22.0&\textbf{22.2}&\textbf{20.9}&\textbf{23.8}&\textbf{25.6}&\textbf{22.69} \\ 
\bottomrule
\end{tabular}
\caption{\textbf{Overall results of our best model and other open-source and closed-source models on the MathVista and MathVision.}
We present the detailed score for all the tasks related to geometry such as ``GPS'' and ``AnaG'', as well as the average score over these tasks in two benchmarks denoted as ``AVG''. Due to limited space, we utilize abbreviations for these geometry-related tasks and illustrate the detailed task name in the Appendix \ref{sec:Detailed_TI}.
\textbf{Bold results} indicate the best results for all models, and the \colorbox{my_blue}{red results} indicate the best results among the open-source models.
\ddag indicates our re-implemented model without an official checkpoint.
$^{*}$ indicates our re-implemented test results missed in benchmarks or origin papers.
InternVL\dag represents the abbreviation for InternVL-Chat-V1.2-Plus.
InternLM-VL\dag represents the abbreviation for InternLM-XComposer2-VL.
The suffix “-G” to the model name indicates a model trained on the GeoGPT4V.
}
\label{tab:appendix_comparison}
\end{table*}

\end{document}